\pdfoutput=1
\documentclass[11pt]{article}

\usepackage{acl}

\usepackage{times}
\usepackage{latexsym}
\usepackage{xspace}
\usepackage{color,xcolor}
\usepackage[T1]{fontenc}
\usepackage[utf8]{inputenc}
\usepackage{amssymb}
\usepackage{utfsym}
\usepackage{pifont}
\usepackage{url}
\usepackage{enumitem}
\usepackage{booktabs}
\usepackage{multirow}
\usepackage{tabularx}
\usepackage{courier}
\usepackage{makecell}
\usepackage{amsmath}
\usepackage[most]{tcolorbox}
\usepackage{algorithm}
\usepackage{algpseudocode}
\usepackage{listings}
\usepackage{xcolor}
\usepackage{graphicx}
\usepackage{fancyvrb}
\usepackage{color}
\usepackage{subcaption}  % 引入subcaption包
\usepackage{wrapfig}
\usepackage{multicol}
\definecolor{color_1}{HTML}{B8D9EB}
\definecolor{color_2}{HTML}{FEFF92}
\definecolor{color_3}{HTML}{99F0C2}

\usepackage{microtype}
\usepackage{inconsolata}

\newcommand{\redtext}[1]{{\color{red}#1}}
\newcommand{\ourbench}[0]{MT-Bench-101\xspace}
\newtcolorbox{tbox}[2][]{colback=yellow!6!white,colframe=blue!35!orange,left=3pt,right=3pt,top=-2pt,bottom=3pt,title=\small{#2},#1}
\title{MT-Bench-101: A Fine-Grained Benchmark for Evaluating Large Language Models in Multi-Turn Dialogues}

\makeatletter
\newcommand{\printfnsymbol}[1]{%
  \textsuperscript{\@fnsymbol{#1}}%
}
\makeatother
\author{
Ge Bai$^{*1}$,
Jie Liu$^{*2,3}$,
Xingyuan Bu$^{*\dag 1}$,
Yancheng He$^{1}$,
Jiaheng Liu$^{1}$,
Zhanhui Zhou$^{3}$, \\
{\bf Zhuoran Lin$^{1}$, Wenbo Su$^{1}$, Tiezheng Ge$^{1}$, Bo Zheng$^{1}$, Wanli Ouyang$^{2,3}$} \\
$^1$Alibaba Group\ \ \ 
$^2$The Chinese University of Hong Kong\\
$^3$Shanghai AI Laboratory \\
{\tt bg427839@taobao.com, xingyuanbu@gmail.com}
}
\begin{document}
\maketitle
\let\thefootnote\relax\footnotetext{$*$ Equal contribution. $\dag$ Corresponding author.}
\begin{abstract}

% Large Language Models (LLMs) have significantly reshaped dialogue systems. However, current benchmarks primarily evaluate these models based on single-turn interactions, overlooking the inherent multi-turn nature of conversations between users and chatbots.
% To address this oversight, we introduce \ourbench,  a new benchmark designed specifically to assess the capabilities of LLMs in multi-turn dialogues.
% By conducting a detailed analysis of real multi-turn dialogue data, we construct a three-tier hierarchical ability taxonomy comprising 4208 dialogue turns across 1388 multi-turn dialogues, allowing for a comprehensive assessment of multi-turn chat abilities. 
% Our extensive analysis of  \needcheck{15} LLMs based on \ourbench identifies GPT-4 as the superior model in multi-turn conversations, highlighting a \needcheck{greater difference between open-source and proprietary models in multi-turn settings compared to single-turn interactions. 
% Moreover, we find that the performance rankings of LLMs on \ourbench significantly differ from those on current benchmarks. }
% Additionally, our study also suggests that common alignment techniques, such as Reinforcement Learning from Human Feedback (RLHF), do not substantially improve multi-turn chat capabilities.

The advent of Large Language Models (LLMs) has drastically enhanced dialogue systems.  However, comprehensively evaluating the dialogue abilities of LLMs remains a challenge. Previous benchmarks have primarily focused on single-turn dialogues or provided coarse-grained and incomplete assessments of multi-turn dialogues, overlooking the complexity and fine-grained nuances of real-life dialogues.
To address this issue, we introduce \ourbench, specifically designed to evaluate the fine-grained abilities of LLMs in multi-turn dialogues. By conducting a detailed analysis of real multi-turn dialogue data, we construct a three-tier hierarchical ability taxonomy comprising 4208 turns across 1388 multi-turn dialogues in 13 distinct tasks. 
We then evaluate 21 popular LLMs based on \ourbench, conducting comprehensive analyses from both ability and task perspectives and observing differing trends in LLMs performance across dialogue turns within various tasks. 
Further analysis indicates that neither utilizing common alignment techniques nor chat-specific designs has led to obvious enhancements in the multi-turn abilities of LLMs.
% , indicating a current deficiency in comprehensive multi-turn dialogue abilities among LLMs.
% Further analysis of models utilizing common alignment techniques and the chat-specific models, reveals no significant improvement in their multi-turn abilities, indicating a current deficiency in comprehensive multi-turn dialogue abilities among LLMs.
% Observing differing trends in LLM performance across dialogue turns underscores the necessity of designing fine-grained ability items. 
Extensive case studies suggest that our designed tasks accurately assess the corresponding multi-turn abilities. The data and code are available at \url{https://github.com/mtbench101/mt-bench-101}.
% , further pointing out the direction for future enhancements in the multi-turn abilities of LLMs.

\end{abstract}

\section{Introduction}

\begin{figure}[t]
    \centering
    \setlength{\abovecaptionskip}{0.15cm}  
    \setlength{\belowcaptionskip}{-0.2cm}
    \includegraphics[width=0.98\linewidth]{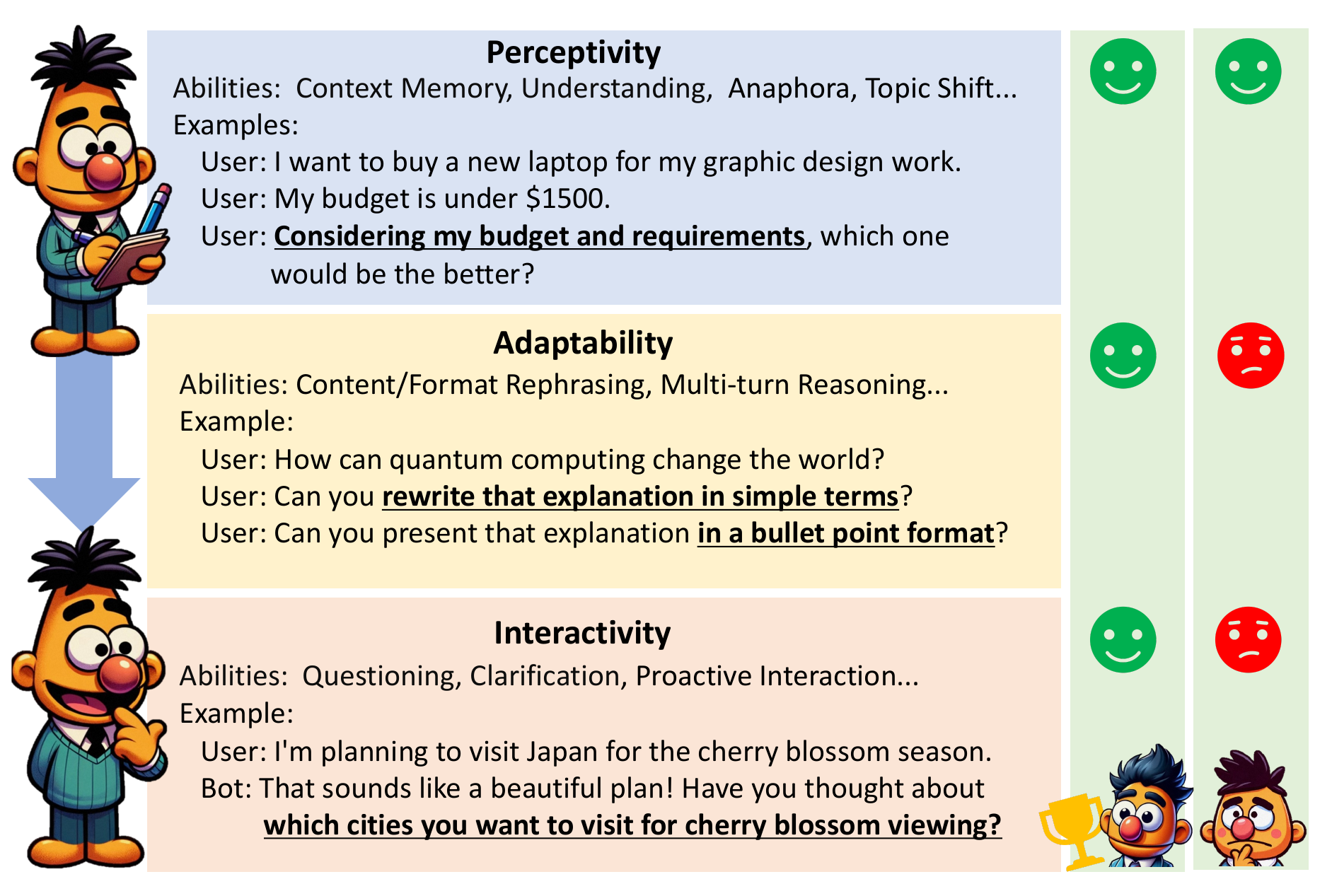}
    \caption{MT-Bench-101 encompasses three overarching abilities and thirteen distinct tasks within multi-turn dialogue scenarios, facilitating a granular benchmarking from basic perceptivity to advanced interactivity. On the right, a model with a broader range of abilities is considered better in multi-turn scenarios.}
    \label{fig: motivation}
    \vspace{-4mm}
\end{figure}

% Large Language Models (LLMs) based chatbots have made remarkable advances and become a game-changing tool in the AI-driven assistant era.
Large Language Models (LLMs) based chatbots have made remarkable advances and significantly enhanced dialogue systems.
Several benchmarks have been introduced to assess the capabilities of Large Language Models (LLMs) in single-turn dialogues, \textit{e.g.}, MMLU~\cite{mmlu}, BBH~\cite{bbh}, and AlpacaEval~\cite{alpaca_eval}. However, daily dialogues between users and chatbots usually involve multi-turn conversations~\cite{sharegpt,realchat}, which include multiple utterances as part of the dialogue history. 
Therefore, it is essential to evaluate the proficiency of LLMs in generating coherent responses utilizing multiple utterances ~\cite{lan2020kind}.
% Unlike single-turn dialogues, multi-turn dialogue models must utilize these multiple utterances to generate coherent responses~\cite{lan2020kind}.
Early studies like MT-bench~\cite{mtbench} mainly focus on two-turn dialogues and coarse-grained abilities, not sufficiently covering the complexity of real-world multi-turn dialogue scenarios.
This indicates a considerable scope for improvement in current benchmarks for multi-turn dialogues, underscoring the urgent need to develop a comprehensive benchmark that can effectively compare the chat abilities of LLMs in multi-turn dialogues.
% The performance gap between open-source and proprietary models is narrowing in some mainstream benchmarks, e.g. MT-bench~\citep{mtbench} and AlpacaEval~\cite{alpacafarm}, with instances where some opensource models even outperform GPT-4. However, the practical usability of open-source models still significantly lags behind their closed-source counterparts in real-world multi-turn dialogues~\citep{botchat}. This phenomenon indicates a fundamental discrepancy between user perceptions of the usefulness of chatbots in multi-turn dialogues and the criteria adopted by current benchmarks.

% We contend that this discrepancy largely stems from existing evaluations, such as AlpacaEval~\citep{dubois2023alpacafarm}, focus primarily on measuring LLMs' core capabilities in single-turn interactions. Some extend this to multi-turn scenarios, like MT-bench, but they do so without fully addressing the unique challenges often encountered in real multi-turn dialogues. To illustrate this, we present a dialogue example with two models addressing an xx question in Figure 1. These models are LLaMA-13B, a base model without fine-tuning, and Vicuna-13B, a variant of LLaMA-13B fine-tuned on high-quality dialogues. Although the Vicuna model outperforms LLaMA in the MT-bench (Table x), its responses in multi-turn scenarios are frequently not preferred by users. 
% This misalignment with current benchmarks underscores an essential requirement: 
% To facilitate the multi-turn dialogue research
In this paper, we introduce \ourbench, a new benchmark designed specifically for evaluating the chat capabilities of LLMs in multi-turn dialogues, as shown in Figure~\ref{fig: motivation}. 
% For ability modeling, we conduct a systematic analysis of real multi-turn dialogue data~\cite{sharegpt, realchat} and the teaching taxonomy of multi-turn dialogues from educational psychology~\cite{alexander2018developing, marchel2007learning}. 
% This has led to the creation of a \textbf{three-tier hierarchical ability taxonomy} that is both data-driven and grounded in psychological principles. 
% %%%%%%%%
During the ability modeling process for multi-turn dialogue, we undertake a systematical analysis combining real-world multi-turn dialogue data~\cite{sharegpt, realchat} with the teaching taxonomy from educational psychology~\cite{alexander2018developing, marchel2007learning}.
This integrated approach has culminated in the formulation of a \textbf{three-tier hierarchical ability taxonomy} which is both data-driven and rooted in psychological frameworks.

\begin{figure}[t]
    \centering
    \includegraphics[width=0.95\linewidth]{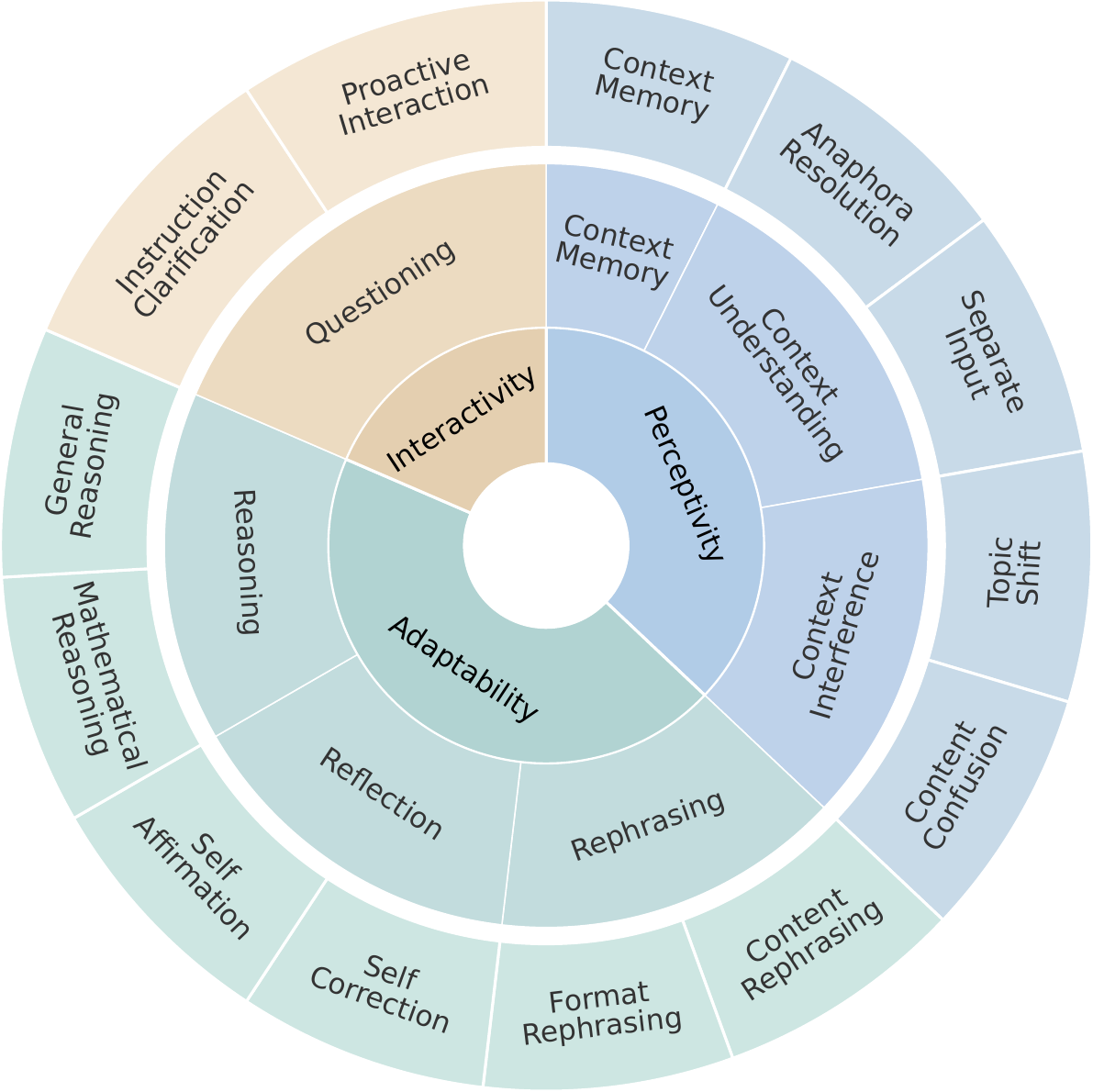}
    \caption{Our three-tier hierarchical ability taxonomy of multi-turn dialogues.}
    \label{fig:diagram-overview}
    \vspace{-0.5cm}
\end{figure}

Figure \ref{fig:diagram-overview} illustrates the overall framework of our ability taxonomy.
The first layer outlines three progressive overarching abilities which are depicted in Figure~\ref{fig: motivation}.
\textbf{Perceptivity} is the most fundamental ability, reflecting the model's accuracy in understanding context. \textbf{Adaptability} is built upon this foundation, indicating the model's ability to respond effectively to user feedback. Finally, \textbf{Interactivity} captures the capacity of models for proactive engagement with humans, which is crucial for excelling in multi-turn interactions.
% As illustrated in Figure ~\ref{fig:diagram-overview}, the first tier of our taxonomy emphasizes that language models must initially exhibit \textbf{Perceptivity} to accurately understand context, followed by \textbf{Adaptability} to effectively respond to user feedback, and \textbf{Interactivity} to proactively engage with humans, which are crucial for excelling in multi-turn interactions. 
The second tier specifies seven detailed abilities, while the third tier further decomposes these abilities into 13 distinct tasks. This taxonomy provides evaluation results across three levels from general to detailed, allowing for the identification of deficiencies in models at varying levels of granularity.
% For data, we choose GPT-4 to generate all samples to avoid data contamination~\cite{sainz2023nlp}. 
% For each third-tier task, we meticulously design specific generation rules and manually craft examples as prompts to generate approximately \needcheck{150} samples utilizing GPT-4. 
For each third-tier task, we meticulously design specific prompts and utilize GPT-4 for data generation.
% Subsequently, we conduct a thorough manual review of all samples, making revisions to enhance the quality of those deemed substandard. 
In total, \ourbench encompasses 4208 turns within 1388 multi-turn dialogues.

% For evaluation, we design a golden-context evaluation system that includes both an overall score and per-turn scores. In most multi-turn dialogues, the user query is posed as a follow-up for the chatbot's response from the preceding turn. Utilizing wrong self-predicted text as dialogue history may make the current turn's query meaningless. To counteract this, we base each turn's context on the dataset's original golden responses for both generation and evaluation. 

For evaluation, we utilize golden context as dialogue history allowing LLMs to generate more fluid and rational dialogues, and we follow~\citealp{mtbench, alpacafarm, botchat} to utilize GPT-4 as a stand-in for human raters to score for each turn.
We design unique scoring guidelines for each task and use the lowest round score as the total score for the dialogue to allow for a more rational assessment.

% we utilize GPT-4 as a stand-in for human raters to score for each turn. We adopt the minimum turn score as the overall score of the dialogue.
% is a more rational approach than the average turn score, as confirmed by human rating experiments in Section~\ref{sec: human_ranking}.

We then perform extensive experiments on \ourbench to assess the multi-turn chat ability of existing LLMs, including 2 close-sourced LLMs and 19 open-sourced LLMs. Our findings include:
% \vspace{-2mm}
\begin{itemize}[leftmargin=4mm]
% We identify adaptability and interactivity as the key deficiencies analyzing from the perspectives of task and capability, and GPT-4 is the most powerful model for multi-turn dialogues. 
% The performance of models within various tasks exhibits differing trends with the progression of turns, reflecting the distinct characteristics of the abilities.
% Model performance improves as the model size increases. However, neither utilizing common alignment techniques nor chat-specific designs has resulted in significant enhancements in the multi-turn abilities of LLMs.

% We analyzed both the task and ability dimensions
    \item We identify adaptability and interactivity as the key deficiencies of existing LLMs, and GPT-4 is the most powerful model for multi-turn dialogues. 
    % \vspace{-2mm}
    \item The average performance of models within various tasks exhibits differing trends with the progression of turns, reflecting the distinct characteristics of the abilities.
    % \vspace{-2mm}
    \item Model performance improves as the model size increases. However, neither utilizing common alignment techniques (such as RLHF) nor chat-specific designs has resulted in significant enhancements in the multi-turn abilities of LLMs.
    % \vspace{-1mm}
    \item The agreement between GPT-4 and human expert evaluations reached 87\%, utilizing our designed evaluation approach.
    % The \ourbench rankings align more closely with the actual ability of models in multi-turn dialogue scenarios.
\end{itemize}

% Notably, our experiments reveal a substantial difference between the rankings of current LLMs on MT-bench and AlpacaEval, and their rankings on our \ourbench multi-turn benchmark (as shown in Figure assets).
% For instance, Yi-13B, which enjoys a higher score on AlpacaEval, scores quite low on \ourbench. This lower score on \ourbench is consistent with Yi-13B's relatively poor performance in actual multi-turn dialogue scenarios. It suggests that \todo.

% We have analyzed real dialogues between humans and chatbots, leveraging data from ShareGPT and RealChat. From this analysis, we propose a three-level hierarchy that categorizes dialogue interactions into three scenarios, seven abilities, and thirteen tasks, as illustrated in Figure~\ref{}

% Our contributions are two-fold:
% \begin{enumerate}
%     \item We categorize the core capabilities required by Large Language Models (LLMs) in multi-turn dialogue scenarios into six categories and develop a comprehensive benchmark, \ourbench.
%     \item We observe that the rankings on our \ourbench benchmark differ significantly from those of current benchmarks, such as MT-bench and AlpacaEval. The \ourbench rankings align more closely with the actual ability of models in multi-turn dialogue scenarios.
%     \item 
% \end{enumerate}

\vspace{-0.35cm}
\section{Related Work}
\vspace{-0.1cm}
\paragraph{LLMs for Multi-turn Dialogues}
Recent advancements in LLMs such as GPT-3.5/GPT-4~\citep{instructgpt,chatgpt,gpt4} have garnered significant attention~\cite{liu20242,wang2023rolellm,feng2022beyond,bu2021gaia,peng2023gaia,unicoder,compress,guo2023owl,li2024graphreader,lyu2024task}. To enhance the multi-turn capabilities of open-sourced LLMs, initial efforts begin with collecting human-ChatGPT dialogues~\citep{sharegpt}, leading to the creation of Vicuna~\citep{vicuna}. RealChat~\citep{realchat} later expands the data to 1 million conversations. To generate more sophisticated datasets, Baize~\citep{Baize} and UltraChat~\cite{UltraChat} employ alternating GPT interactions. Parrot~\citep{parrot} trains a model to simulate the user, thereby generating improved data. Further, Cue-CoT and ICL-AIF~\citep{cuecot,iclaif} enhance model capabilities for multi-turn interactions through In-Context-Learning (ICL)~\citep{gpt3} and Chain-of-Thought (CoT)~\citep{cot} algorithms.

\vspace{-0.12cm}

\paragraph{Benchmarks for Multi-turn LLMs}
Most benchmarks evaluate LLMs through single-turn instructions~\citep{mmlu}, missing the nuance of human conversation. To investigate multi-turn ability, ABC-Eval~\citep{abceval} relies on labor-intensive human evaluations. AlpacaEval~\citep{alpaca_eval} and PandaLM~\citep{pandalm} attempt to automatically assess open-ended instructions, but they remain confined to single-turn settings. MT-Bench~\citep{mtbench} and MT-Bench++~\citep{parrot} expand multi-turn evaluations across eight topics. 
% MT-Eval~\citep{mteval} further evaluates LLMs on four tasks. 
BotChat~\citep{botchat} and MINT~\citep{mint} focus on specialized tasks such as dialogue generation abilities. Despite these efforts, there remains a notable gap in fine-grained evaluations for multi-turn interactions.

\begin{table*}[h]
\centering
\setlength{\abovecaptionskip}{0.15cm}  
\setlength{\belowcaptionskip}{-0.3cm}  
\resizebox{1\textwidth}{!}{
\begin{tabular}{l|c|l}
\toprule
\textbf{Task} & \textbf{Abbr.} & \textbf{Description}  \\ \midrule
Context Memory & CM & Recall early dialogue details to address the user's current question. \\ \midrule
Anaphora Resolution & AR & Identify pronoun referents throughout a multi-turn dialogue. \\ 
Separate Input & SI & The first turn outlines the task requirements and the following turns specify the task input. \\ \midrule
Topic Shift & TS & Recognize and focus on the new topic when users unpredictably switch topics. \\ 
Content Confusion & CC & Avoid interference from similar-looking queries with distinct meanings in the dialogue's history. \\ \midrule
Content Rephrasing & CR & Rephrase the content of the last response according to the user's newest requirement. \\ 
Format Rephrasing & FR & Rephrase the format of the last response according to the user's newest requirement. \\ \midrule
Self-correction & SC & Recorrect the last response according to the user feedback. \\ 
Self-affirmation & SA & Preserve the last response against inaccurate user feedback. \\ \midrule
Mathematical Reasoning & MR & Collaboratively solve complex mathematical problems with users across dialogue turns.\\ 
General Reasoning & GR & Collaboratively solve complex general reasoning problems with users across dialogue turns. \\ \midrule
Instruction Clarification & IC & Seek clarification by asking further questions on ambiguous user queries. \\
Proactive Interaction & PI & Propose questions in reaction to user statements to spark their interest to continue the dialogue. \\
\bottomrule
\end{tabular}
}
% \vspace{-1mm}
\caption{The 13 tasks for multi-turn dialogues within \ourbench.}
\label{tab:task}
\vspace{-2mm}
\end{table*}

\vspace{-0.15cm}
\paragraph{Benchmarks for Fine-grained Abilities}
The advent of general LLMs has highlighted the need for more comprehensive evaluation. ~\citealp{mmlu} introduces MMLU with an extensive suite of 57 tasks spanning social science and STEM.
Subsequent benchmarks~\citep{ceval, cmmlu, zhong2023agieval, bbh, kola, li2024cif, guo2024codeeditorbench} aim to rigorously assess LLMs' knowledge and logic. ConceptMath~\citep{wu2024conceptmath} and FollowBench~\citep{jiang2023followbench} develop a hierarchical framework for evaluating model capabilities.
\vspace{-0.15cm}
\section{\ourbench}
\vspace{-0.15cm}

This section begins with a detailed description of the three-tier hierarchical ability taxonomy for multi-turn dialogues. Following that, we explain the methodology used to collect the dataset. Finally, we present an analysis of the dataset's statistics.

\vspace{-0.15cm}
\subsection{Hierarchical Ability Taxonomy}
\vspace{-0.12cm}

After analyzing real dialogues from ShareGPT \cite{sharegpt} and RealChat \cite{realchat} and the teaching taxonomy of multi-turn dialogues from educational psychology \cite{alexander2018developing, marchel2007learning, peng2020large, gao2018solution}, we have developed a hierarchical taxonomy of abilities crucial for chatbots to engage effectively in multi-turn dialogues with human users. This taxonomy is structured into three levels, with the third level encompassing 13 distinct tasks. Table \ref{tab:task} provides a brief one-sentence description for each third-level task. This section will deliver a detailed explanation of these three-level abilities and tasks. 
We also provide cases for each task in the Appendix \ref{app: case}.

\vspace{-0.12cm}
\subsubsection{Perceptivity}

Perceptivity requires chatbots to adeptly track and use historical dialogue data to provide logical and consistent responses, encompassing the following three core abilities.

\vspace{-0.12cm}
\paragraph{Context Memory:}
To ensure continuity and relevance in dialogues, chatbots must exhibit a robust context memory capability. This involves accurately retrieving and utilizing past dialogue information to address current user inquiries. We named \textit{Context Memory} (CM) for the third-level ability.
% At the beginning of a conversation, users might share specific details. The model must recall and effectively use this information in later stages of the dialogue to provide pertinent responses. 
% An illustration is provided in Figure~\ref{fig: turns_self}.

% This capability emphasizes the chatbot's understanding of the dialogue across multiple turns, for which we propose two tasks:
\vspace{-0.12cm}
\paragraph{Context Understanding:}
\textit{Anaphora Resolution} (AR). It is common for users to use demonstrative pronouns (e.g., "these," "it") in dialogues. A key ability for chatbots is to identify the referents of these pronouns accurately to generate appropriate responses;
\textit{Separate Input} (SI).
Dialogues typically unfold over several turns, with the initial turn outlining the task requirements and subsequent turns specifying the task input. Understanding the relationship between instructions and inputs is demanded for effective chatbots.

\vspace{-0.12cm}
\paragraph{Context Interference:}
\textit{Topic Shift} (TS).
Users might unpredictively switch topics in multi-turn dialogues. This task assesses the chatbot's ability to recognize a topic shift and ignore unrelated preceding information, thereby concentrating on the new topic at hand;
\textit{Content Confusion} (CC) centers on the chatbot's skill in managing situations where users pose questions that, while textually similar in history questions, necessitate distinct responses.
% , as demonstrated in Figure~\ref{fig: case_similar}.
% For example, a user might discuss "Apple Computers" in one turn and then switch to asking whether "apples can help with weight loss" in the next.

\vspace{-0.12cm}
\subsubsection{Adaptability}
\vspace{-0.12cm}
Chatbots adjust their early responses with the user's new requirements (Rephrasing), new conditions, and hypotheses (Reasoning) and can correct or insist on answers according to user-challenging feedback (Reflective) in user-triggered dialogue.

\vspace{-0.12cm}
\paragraph{Rephrasing:}
\textit{Content Rephrasing} (CR) requires chatbots to have a thorough understanding of the text to rephrase the content of the last response based on the user’s latest requirement (e.g., summarizing this paragraph). 
\textit{Format Rephrasing} (FR) involves a transformation in structure while preserving the original information (e.g., converting this paragraph into a list format).

\vspace{-0.12cm}
\paragraph{Reflection:}
\textit{Self-correction} (SC).
Upon receiving user feedback indicating skepticism or errors in the last response, the chatbot will correct mistakes, and provide a more accurate subsequent response. 
Within our dataset, this task is limited to instances where the chatbot's initial reply was erroneous or not precise, and the user's critique is deemed valid.
% This task entails the chatbot reviewing its last message, correcting any mistakes, and providing a more accurate subsequent response.
\textit{Self-affirmation} (SA).
Unlike self-correction task, self-affirmation comes into play when the chatbot's initial response is correct or accurate, yet it encounters incorrect feedback from the user. In such cases, the chatbot needs to identify the inaccuracies in the user's feedback and adhere to its original response. 

\vspace{-0.13cm}
\paragraph{Reasoning:}
\textit{Mathematical Reasoning} (MR). Effective reasoning across multiple dialogue turns is essential in solving mathematical problems, as users may introduce new conditions or hypotheses as the conversation progresses.
\textit{General Reasoning} (GR) encompasses a variety of reasoning challenges, e.g. puzzles, inductive reasoning, and deductive reasoning. Chatbots are required to work alongside users through successive dialogue turns to address these issues.
% , further detailed in Table~\ref{}.
\vspace{-0.1cm}
\subsubsection{Interactivity}
Chatbots proactively propose questions to guide the dialogue or gather information for better responses in chatbot-triggered dialogue.

\vspace{-0.13cm}
\paragraph{Questioning:}
\textit{Instruction Clarification} (IC) targets scenarios where the user's initial question is unclear. The chatbot needs to ask follow-up questions to obtain more information. This iterative intent clarification process may span several turns to ensure the chatbot fully grasps the user's intent.
\textit{Proactive Interaction} (PI) assesses the chatbot's ability to craft suitable follow-up questions or comments in reaction to user statements, thereby sparking the user's interest to continue the dialogue.

\vspace{-0.15cm}
\subsection{Data Collection}
\vspace{-0.05cm}
% Datasets from traditional tasks (e.g. mutual\cite{}) are \todo{}.
% Following~\ref{}, we employ GPT-4 to produce our dataset. This also prevents data leakage~\cite{}. For each task, we elaborately design a prompt to generate the dialogues. Appendix~\ref{} shows all prompts used for generation. \ourbench covers A wealth of subjects: \todo{}.
\begin{table}[t]
\centering
\setlength{\abovecaptionskip}{0.15cm}  
\setlength{\belowcaptionskip}{-0.3cm}  
\resizebox{0.48\textwidth}{!}{
\begin{tabular}{ccccc}
\toprule
Benchmark & \#Dialogues & \#Turns & \#Tasks & Fine-grained \\
\midrule
AlpacaEval & 805 & 805 & 1 & \ding{55} \\
MT-Bench & 80 & 160 & 1 & \ding{55} \\
MT-Bench++ & 80 & 640 & 1 & \ding{55} \\
BotChat & 547 & 547 & 1 & \ding{55} \\
MINT & 568 & 568 & 3 & \ding{55} \\
% MT-Eval & 168 & 1170 & 4 & \ding{55} \\
\midrule
\textbf{\ourbench} & \textbf{1388} & \textbf{4208} & \textbf{13} & \ding{51} \\
\bottomrule
\end{tabular}
}
\caption{Data statistics.}
\label{table:datacomparison}
\vspace{-4mm}
\end{table}
We tailored unique data generation prompts for each task based on its specific characteristics and utilized GPT-4 to construct data. In detail, the prompts included data generation rules and used manually crafted examples as guidance for GPT-4. This ensured the generated data met the specific needs of each task. Our benchmark covers 30 diverse topics, including health, history, science, finance, law, humanities, arts, and others. The Appendix~\ref{app: data_gener} shows a complete list of topics and the prompts used for data generation.

For each task, we utilized GPT-4 to generate over 1000 samples. These samples were then rigorously curated by human annotators to form the final dataset. For each piece of data, it underwent screening by five annotators, and we ultimately retained only the data that all annotators deemed to be of high quality. The primary criteria for curation are shown in the Appendix~\ref{app: data_gener}.

% After generating preliminary data using GPT-4, we manually filtered the data based on difficulty and diversity and removed any offensive and identifying information, resulting in our evaluation dataset. 
% It is important to note that the collection processes were conducted by a team of five members, each holding a university degree in an engineering discipline.

% \vspace{-0.25cm}

% \begin{table}[h]
% \centering
% \resizebox{0.48\textwidth}{!}{
% \begin{tabular}{ccccc}
% \toprule
% Benchmark   & \# Avg. Turns & \#Topics & \#Size & \#Finegrained \\ \midrule
% AlpacaEval & 1             &          & 805    &               \\
% MT-Bench    & 1.5           &          & 80     &               \\
% MT-Bench++  & 8             &          & 80     &               \\
% BotChat     & 6.6           &          & 547    &               \\ \midrule
% Ours        &      3.03         &     30     &        &               \\ \bottomrule
% \end{tabular}
% }
% \caption{Data statistics.}
% \end{table}

\vspace{-0.25cm}
\subsection{Data Statistics}
\vspace{-0.12cm}
Table~\ref{table:datacomparison} shows the key statistics of our \ourbench. This benchmark features a comprehensive hierarchical taxonomy for multi-turn dialogues with 13 distinct tasks, 1388 dialogues, and 4208 turns. Detailed statistics for each task can be found in the Appendix~\ref{app: datastat}.
Additionally, we provide a comparative analysis between \ourbench and existing dialogue evaluation benchmarks. This comparison highlights that \ourbench is the first dataset to specifically focus on fine-grained multi-turn dialogue abilities, notable for its extensive volume of data and diversity of tasks.

% Contextual Memory, Anaphora Resolution, Separate Input, Contextual Mathematic, General Reasoning, Content rephrasing, Format rephrasing, 
%  Self-correction, Self-affirmation,Topic Switch, Content Confusion, Instruction Clarification, Proactive Interaction
\vspace{-0.15cm}
\subsection{Evaluation}
\vspace{-0.15cm}
\label{sec: eval}
% In light of the characteristics of multi-turn dialogues in our benchmark, we have devised a suitable evaluation method. Specifically, 
In multi-turn dialogues, new turns rely on the interaction between humans and chatbots in the preceding turns. This phenomenon is especially significant in tasks with strong interactivity such as instruction clarification and proactive interaction. Hence, we leverage our meticulously curated dataset as the golden context for dialogue history, as opposed to relying on self-predicted context from LLM subjects. This approach facilitates the creation of smoother, more rational dialogues. Moreover, evaluating only the newest response of the LLMs while maintaining consistency with the conversation history also promotes fair evaluation.

\begin{table*}[t]
\centering
\setlength\tabcolsep{5pt}
\renewcommand{\arraystretch}{1.3}
\resizebox{1\linewidth}{!}{
\small
\begin{tabular}{cc|c|cc|cc|cc|cc|cc|cc}
\toprule
\multirow{3}{*}{\textbf{Model}} & \multicolumn{1}{c}{\textbf{}} & \multicolumn{5}{|c}{\textbf{Perceptivity}} & \multicolumn{6}{|c}{\textbf{Adaptability}} & \multicolumn{2}{|c}{\textbf{Interactivity}} \\
 & \multicolumn{1}{c}{\textbf{}} & \multicolumn{1}{|c}{\textbf{Memory}} & \multicolumn{2}{|c}{\textbf{Understanding}} & \multicolumn{2}{|c}{\textbf{Interference}} & \multicolumn{2}{|c}{\textbf{Rephrasing}} & \multicolumn{2}{|c}{\textbf{Reflection}} & \multicolumn{2}{|c}{\textbf{Reasoning}} & \multicolumn{2}{|c}{\textbf{Questioning}} \\
 & \multicolumn{1}{c}{\textbf{Avg.}} & \multicolumn{1}{|c}{\textbf{CM}}     & \multicolumn{1}{|c}{\textbf{SI}} & \multicolumn{1}{c}{\textbf{AR}} & \multicolumn{1}{|c}{\textbf{TS}} & \multicolumn{1}{c}{\textbf{CC}} & \multicolumn{1}{|c}{\textbf{CR}} & \multicolumn{1}{c}{\textbf{FR}} & \multicolumn{1}{|c}{\textbf{SC}} & \multicolumn{1}{c}{\textbf{SA}} & \multicolumn{1}{|c}{\textbf{MR}} & \multicolumn{1}{c}{\textbf{GR}} & \multicolumn{1}{|c}{\textbf{IC}} & \multicolumn{1}{c}{\textbf{PI}} \\
\midrule
Llama2-7B-Chat                  & 6.53                             & 7.64                                & 6.21                            & 7.92                            & 8.23                            & 8.50                            & 8.32                            & 8.56                            & 8.45                            & 4.97                            & 1.88                            & 3.83                            & 5.23                            & 5.11                            \\
Qwen-7B-Chat                     & 7.12                             & 7.65                                & 7.75                            & 8.73                            & 8.42                            & 8.76                            & 8.89                            & 9.16                            & 8.49                            & 7.28                            & 2.25                            & 3.57                            & 5.41                            & 6.24                            \\
ChatGLM2-6B                      & 5.56                             & 6.14                                & 4.69                            & 7.27                            & 6.13                            & 6.26                            & 7.47                            & 7.98                            & 6.97                            & 4.19                            & 2.11                            & 3.00                            & 5.16                            & 4.90                            \\
ChatGLM3-6B                      & 6.47                             & 7.16                                & 5.42                            & 8.21                            & 7.43                            & 8.03                            & 8.38                            & 8.81                            & 7.40                            & 5.63                            & 2.60                            & 3.21                            & 6.19                            & 5.61                            \\
InternLM2-Chat-7B-SFT            & 6.69                             & 7.51                                & 6.26                            & 8.01                            & 8.06                            & 8.70                            & 8.50                            & 8.50                            & 7.68                            & 6.16                            & 3.47                            & 4.48                            & 4.92                            & 4.76                            \\
Yi-6B-Chat                       & 6.93                             & 7.57                                & 5.27                            & 8.69                            & 8.37                            & 8.76                            & 8.43                            & 8.44                            & 7.49                            & 7.85                            & 2.18                            & 3.80                            & 7.30                            & 6.00                            \\
Mistral-7B-Instruct-v0.2         & 6.95                             & 7.66                                & 5.64                            & 8.09                            & 8.30                            & 9.35                            & 8.69                            & 8.59                            & 8.16                            & 7.33                            & 2.58                            & 4.52                            & 5.80                            & 5.66                            \\
\midrule
Vicuna-13B-v1.5                  & 6.37                             & 7.06                                & 5.62                            & 7.81                            & 7.45                            & 8.79                            & 7.96                            & 7.72                            & 7.47                            & 6.70                            & 2.31                            & 4.03                            & 5.05                            & 4.80                            \\
Baize-13B-v2 & 6.12 & 6.78 & 5.15 & 7.86 & 7.40 & 8.07 & 7.96 & 8.15 & 7.24 & 6.32 & 1.67 & 3.69 & 4.35 & 4.95 \\
UltraLM-13B-v2.0 & 4.61 & 4.66 & 4.89 & 5.99 & 6.49 & 8.48 & 2.87 & 2.53 & 6.70 & 5.27 & 1.46 & 2.34 & 4.13 & 4.11 \\
Llama2-13B-Chat                 & 7.15                             & 8.03                                & 7.11                            & 9.00                            & 9.39                            & 8.81                            & 9.07                            & 9.11                            & 7.63                            & 7.60                            & 1.75                            & 3.16                            & 6.07                            & 6.23                            \\
Qwen-14B-Chat                    & 7.82                             & 8.33                                & 8.36                            & 9.04                            & 9.22                            & 9.50                            & 9.12                            & 9.39                            & 8.41                            & 7.97                            & 3.50                            & 4.55                            & 8.21                            & 6.12                            \\
Baichuan2-13B-Chat               & 7.00                             & 7.71                                & 6.38                            & 8.92                            & 8.36                            & 9.07                            & 9.10                            & 8.95                            & 7.75                            & 6.57                            & 2.50                            & 3.65                            & 6.95                            & 5.15                            \\
InternLM2-Chat-20B-SFT           & 6.95                             & 7.35                                & 6.44                            & 8.08                            & 8.05                            & 9.10                            & 8.59                            & 8.55                            & 7.62                            & 7.36                            & 4.05                            & 5.24                            & 4.99                            & 4.99                            \\
Yi-34B-Chat                      & 8.10                             & 8.55                                & 6.79                            & 9.34                            & 9.84                            & 9.34                            & 9.08                            & 9.38                            & 9.01                            & 9.04                            & 4.07                            & 5.90                            & 8.51                            & 6.39                            \\
Mixtral-8x7B-Instruct-v0.1       & 7.38                             & 7.86                                & 5.94                            & 8.49                            & 9.01                            & 9.52                            & 8.91                            & 9.01                            & 8.69                            & 7.78                            & 4.19                            & 5.14                            & 6.03                            & 5.36                            \\
\midrule
GPT-3.5                          & 7.99                             & 8.77                                & 7.67                            & 7.67                            & 9.68                            & 9.87                            & 9.56                            & 9.51                            & 9.18                            & 7.23                            & 4.48                            & 5.31                            & 8.57                            & 6.32                            \\
GPT-4 & 8.86 & 8.88 & 8.99 & 9.58 & 9.83 & 9.98 & 9.54 & 9.57 & 9.36 & 9.52 & 7.15 & 7.17 & 9.00 & 6.64 \\
\midrule
Avg. & 6.92 & 7.52 & 6.37 & 8.26 & 7.72 & 8.24 & 8.36 & 8.44 & 7.98 & 6.93 & 3.61 & 4.84 & 6.22 & 5.52 \\
\bottomrule
\end{tabular}}
\vspace{-1mm}
\caption{The performance of different LLMs on the 13 multi-turn dialogue tasks in our \ourbench. Due to space constraints, the 13 tasks are represented by their corresponding acronyms.}
\label{table:main_result}
\vspace{-2mm}
\end{table*}

\begin{figure*}[t]
    \centering
    % 第一行的两个子图形
    \begin{subfigure}[b]{0.32\linewidth}
        \includegraphics[width=\textwidth]{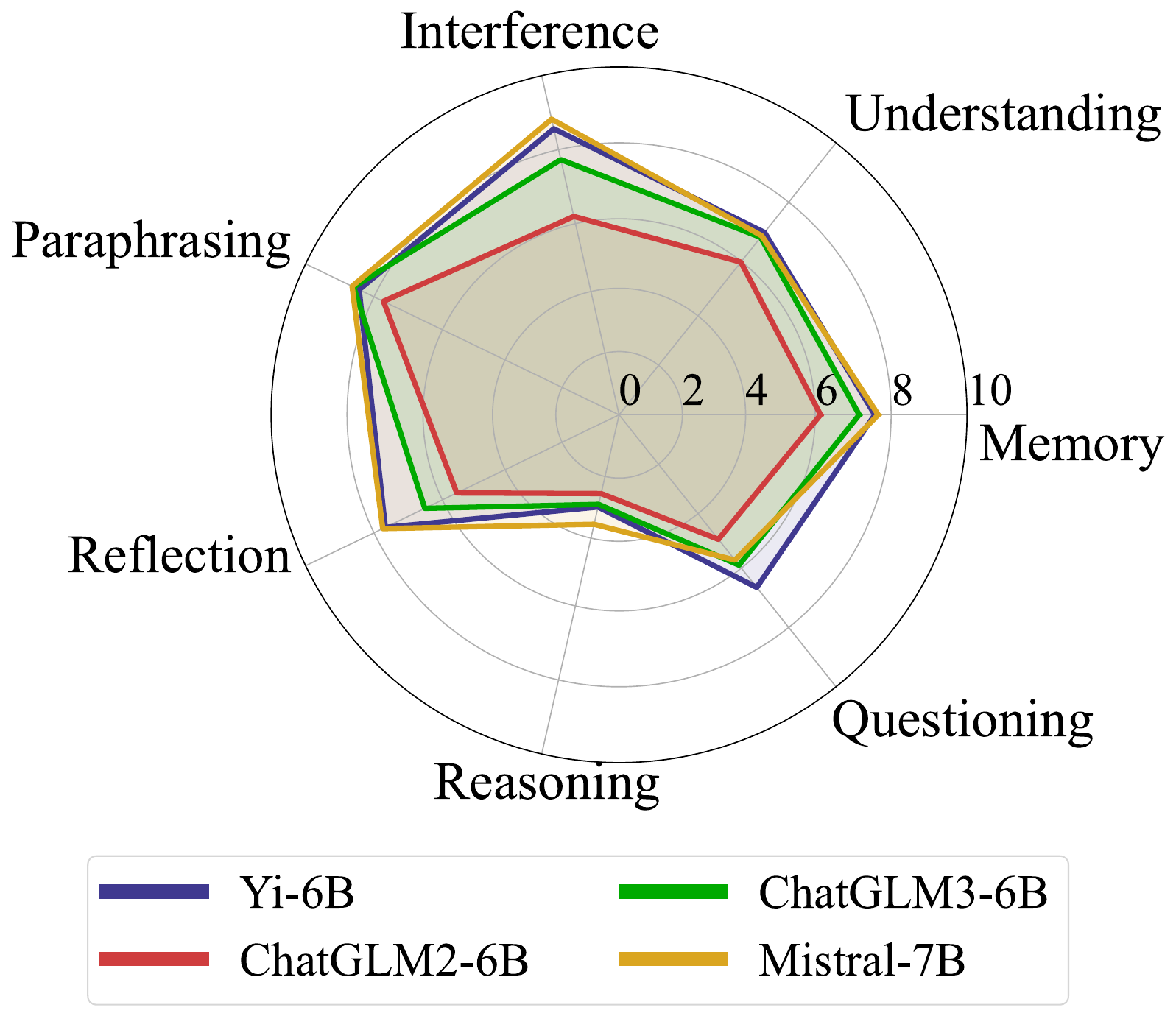}
        % \caption{}
        \label{fig:sub0_1}
    \end{subfigure}
    % \hfill  % 用 hfill 来填充两个子图形之间的空间
    \begin{subfigure}[b]{0.32\linewidth}
        \includegraphics[width=\textwidth]{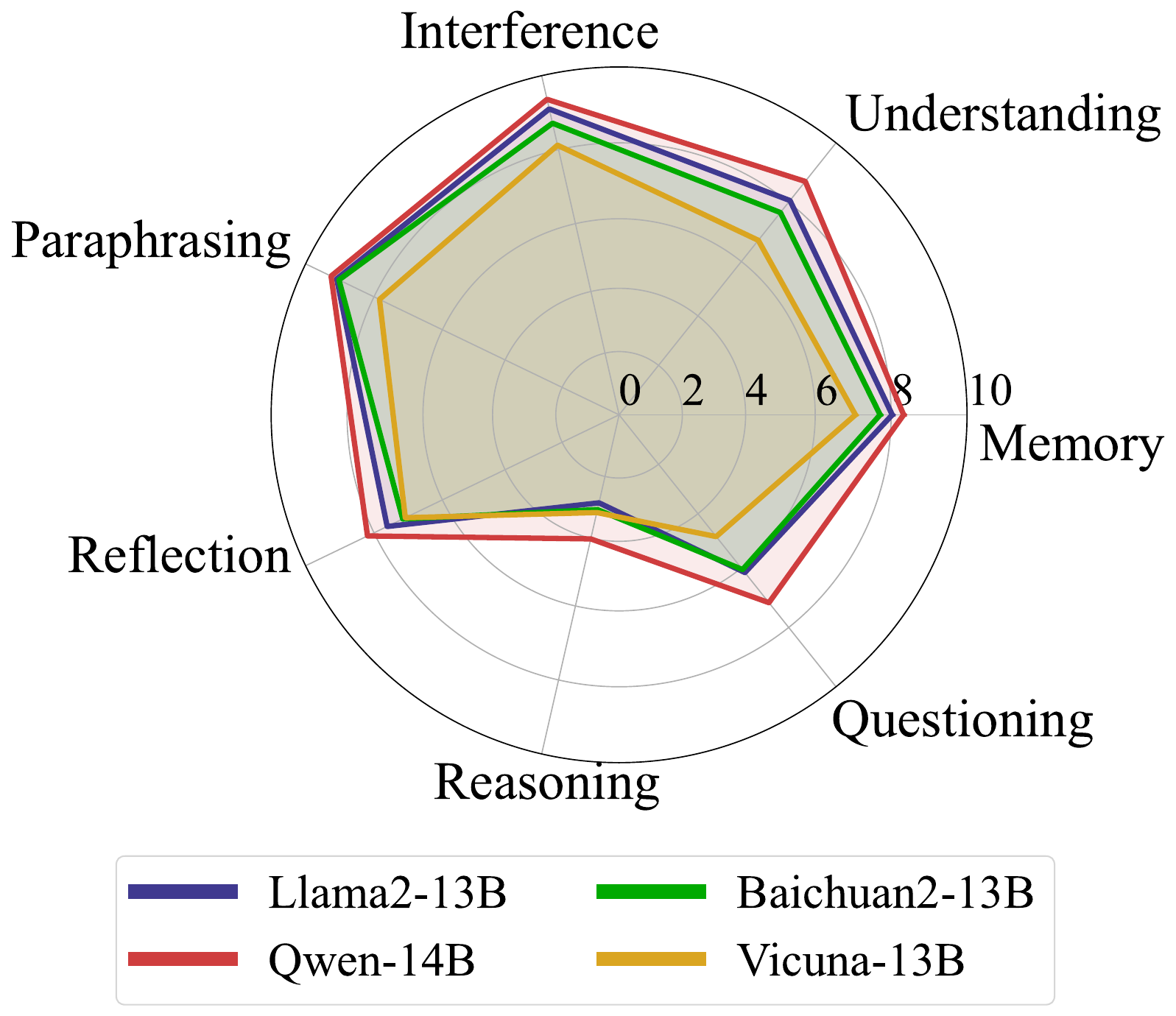}
        % \caption{}
        \label{fig:sub0_2}
    \end{subfigure}
    % \hfill  % 用 hfill 来填充两个子图形之间的空间
    \begin{subfigure}[b]{0.32\linewidth}
        \includegraphics[width=\textwidth]{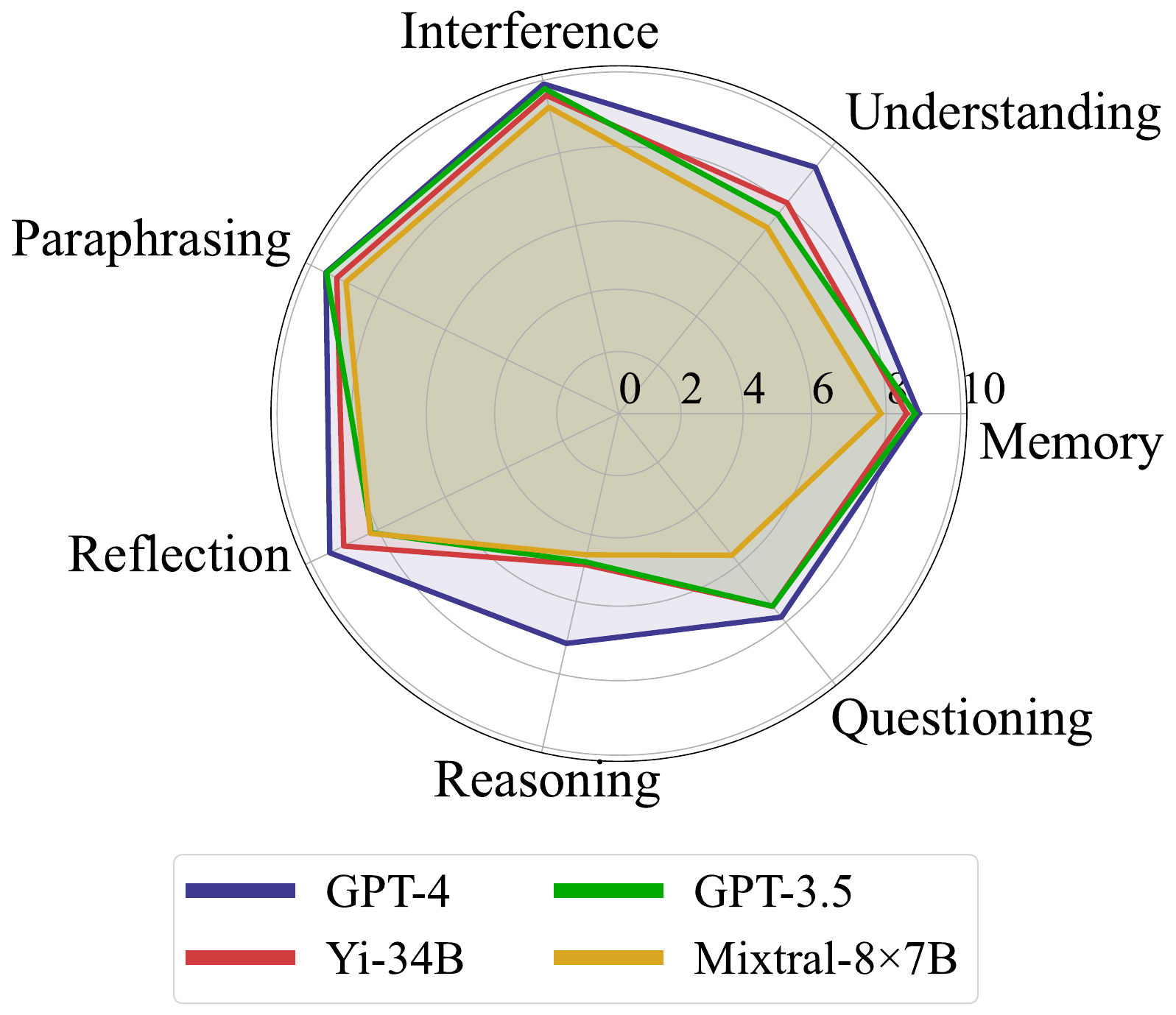}
        % \caption{}
        \label{fig:sub0_3}
    \end{subfigure}
    \vspace{-4mm}
    \caption{Performance of various LLMs for each ability dimension.}
    \label{fig: main_results}
    \vspace{-4mm}
\end{figure*}

Following MT-Bench~\citep{mtbench}, we employ GPT-4 for evaluation in our benchmark. 
% While maintaining a consistent zero-shot prompt framework, 
We tailor different evaluation prompts (see Appendix \ref{app: eval}) for each task and develop fine-grained scoring guidelines detailing what is required for each score level or grade. Then GPT-4 scores each turn of the chatbot's responses from 1 to 10 and gives detailed justifications.
Additionally, our evaluation process utilizes a minimum-score-taking metric, where the lowest score of a turn is considered the final score for the entire dialogue. This approach is consistent with human intuition, as discussed in section~\ref{sec: human_ranking}, because a single failed response can compromise the entire dialogue in closely related conversational contexts. Moreover, this metric prevents models from achieving inflated scores by simply learning patterns from the golden context. This phenomenon will be further explored in section \ref{sec: turns}. 

\citealt{arenahard2024,he2022blind} point out that there is self-bias in LLM judges (e.g., GPT-4 Judge prefers GPT-4 answers). We also provide a leaderboard with Qwen-72B-Chat as the judge model in the Appendix~\ref{app: qwen}, showing that this problem is minor in our benchmark, with the rankings of GPT-4-Judge and Qwen-72B-Judge being consistent.

\vspace{-0.2cm}
\section{Experiments}
\vspace{-0.2cm}

\subsection{Experimental Setup}
\vspace{-0.05cm}
\paragraph{Settings}
We utilize the golden contexts as dialogue histories in all experiments unless otherwise specified. For each LLM, we apply the corresponding chat format and the system prompt while setting the temperature and sampling parameter as their official configures. As for LLM-Judge(GPT-4 and Qwen-72B-Chat), we set the temperature to 0.6. Additional details on the experimental setup and implementation can be found in the Appendix~\ref{app: eval}.
% It's noteworthy that in format rephrasing, content rephrasing, anaphora resolution, self-affirmation, self-correction, and context memory tasks, models generate responses directly from the second turn of dialogue, utilizing the golden context from the first turn as historical dialogue information. (See Appendix~\ref{app: xxx} for more details)
% This approach is adopted due to the task characteristics, where assessment of the respective abilities begins from the second turn of dialogue. While models need to utilize the dialogue history from the first turn in subsequent turns, directly responding to the content of the first turn lacks practical significance.
\vspace{-0.15cm}
\paragraph{Models}
We evaluate 21 popular LLMs on \ourbench, including 2 close-sourced LLMs (\textit{i.e.}, GPT-3.5/ GPT-4~\citep{gpt4}) and 19 open-sourced LLMs (\textit{i.e.}, Llama2-Chat (7B, 13B)~\citep{llama2}, Mistral-Instruct (7B, 8x7B, DPO)~\citep{jiang2023mistral}, Qwen-Chat (7B, 14B)~\citep{qwen}, Yi-Chat (6B, 34B)~\citep{yi}, ChatGLM2-6B/ChatGLM3-6B~\citep{glm}, InternLM2-Chat (7B, 20B, RLHF)~\citep{2023internlm}, Vicuna-13B-v1.5~\citep{vicuna}, Baichuan2-Chat-13B \citep{baichuan}), UltraLM-13B-v2.0~\citep{UltraChat}, and Baize-v2-13B~\citep{Baize}.
More details of these evaluated models can be seen in the Appendix~\ref{app: em}.

\begin{figure}[t]
    \centering
    % 第一行的两个子图形
    \begin{subfigure}[b]{0.49\linewidth}
        \includegraphics[width=\textwidth]{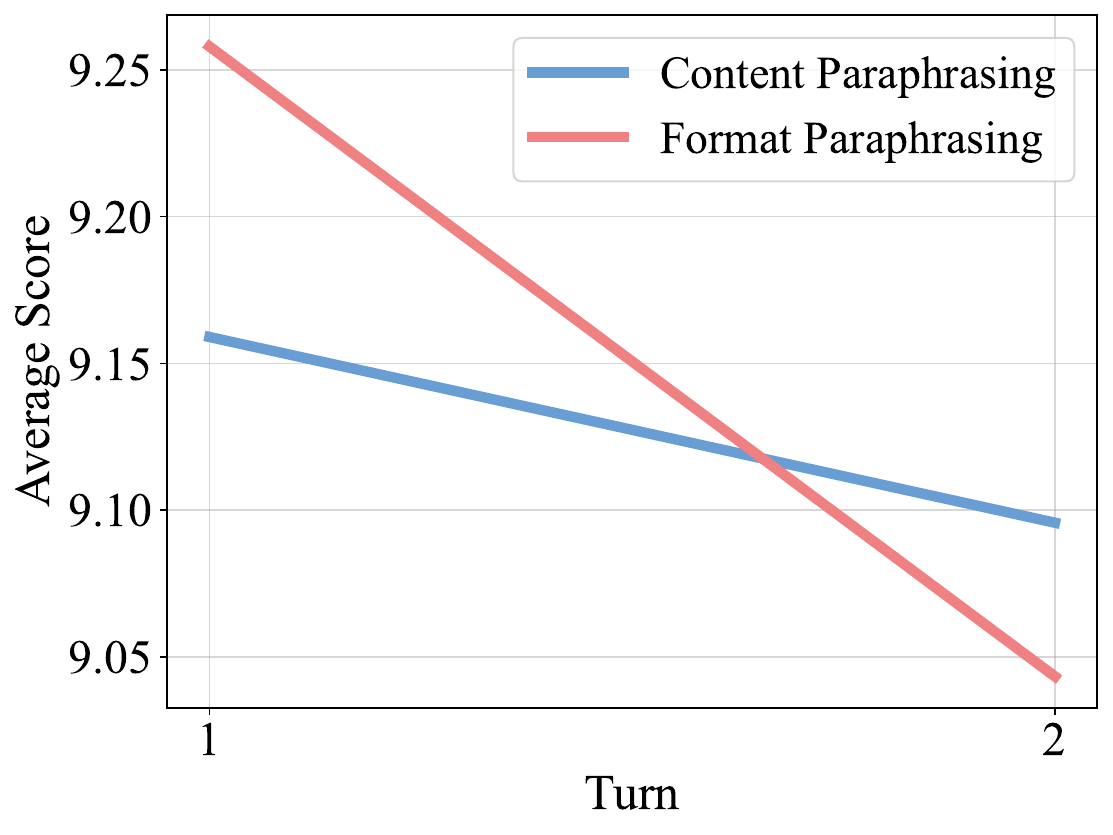}
        \caption{}
        \label{fig:sub1}
    \end{subfigure}
    % \hfill  % 用 hfill 来填充两个子图形之间的空间
    \begin{subfigure}[b]{0.49\linewidth}
        \includegraphics[width=\textwidth]{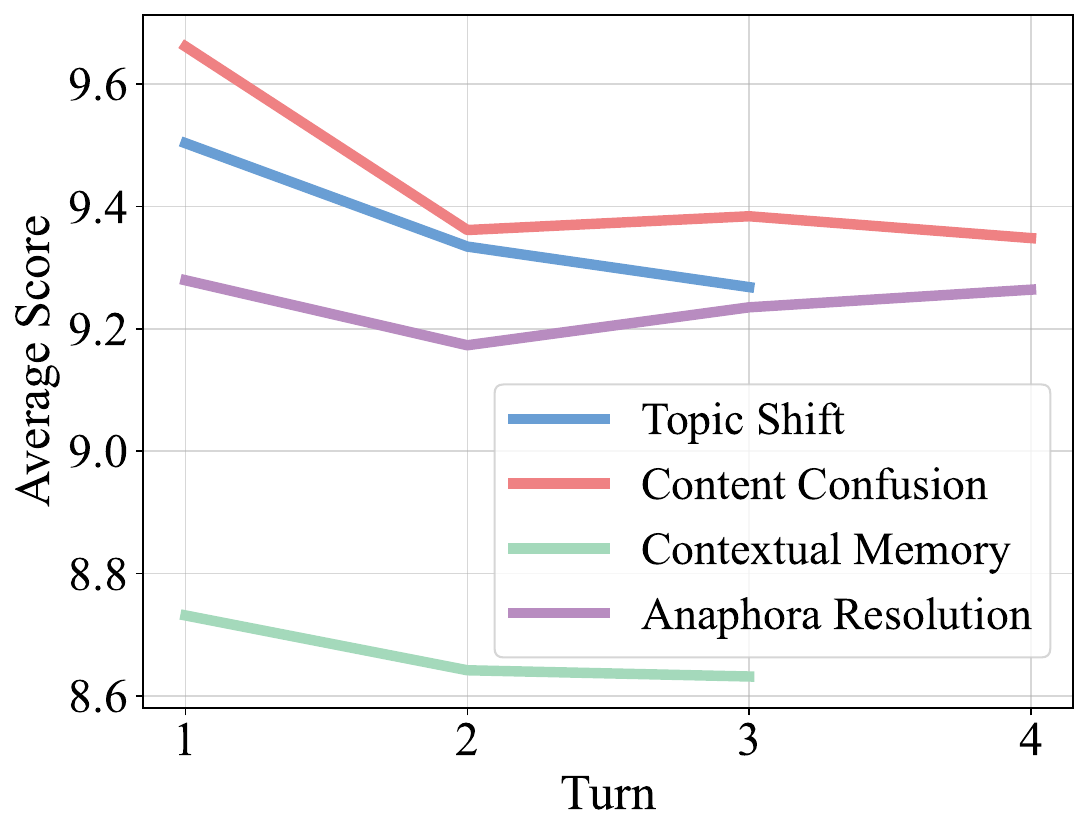}
        \caption{}
        \label{fig:sub2}
    \end{subfigure}
    
    %\vspace{1cm}  % 第一行和第二行之间的垂直空间
    
    % 第二行的两个子图形
    \begin{subfigure}[b]{0.49\linewidth}
        \includegraphics[width=\textwidth]{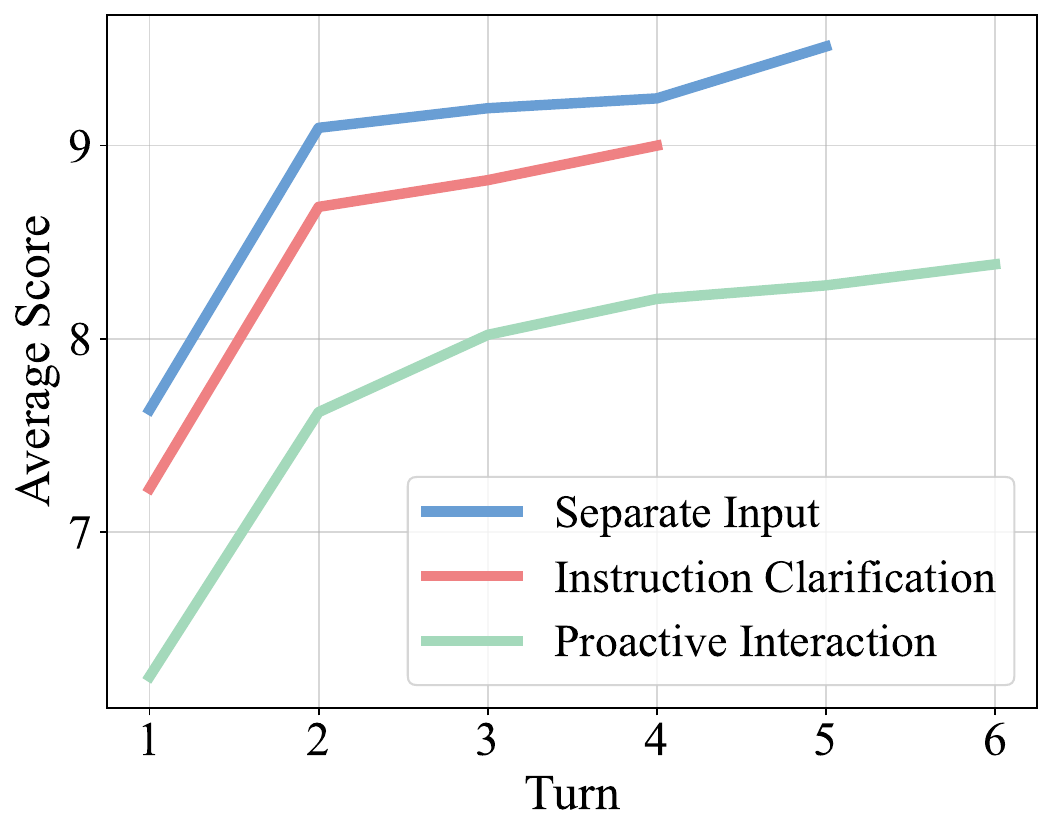}
        \caption{}
        \label{fig:sub3}
    \end{subfigure}
    % \hfill
    \begin{subfigure}[b]{0.49\linewidth}
        \includegraphics[width=\textwidth]{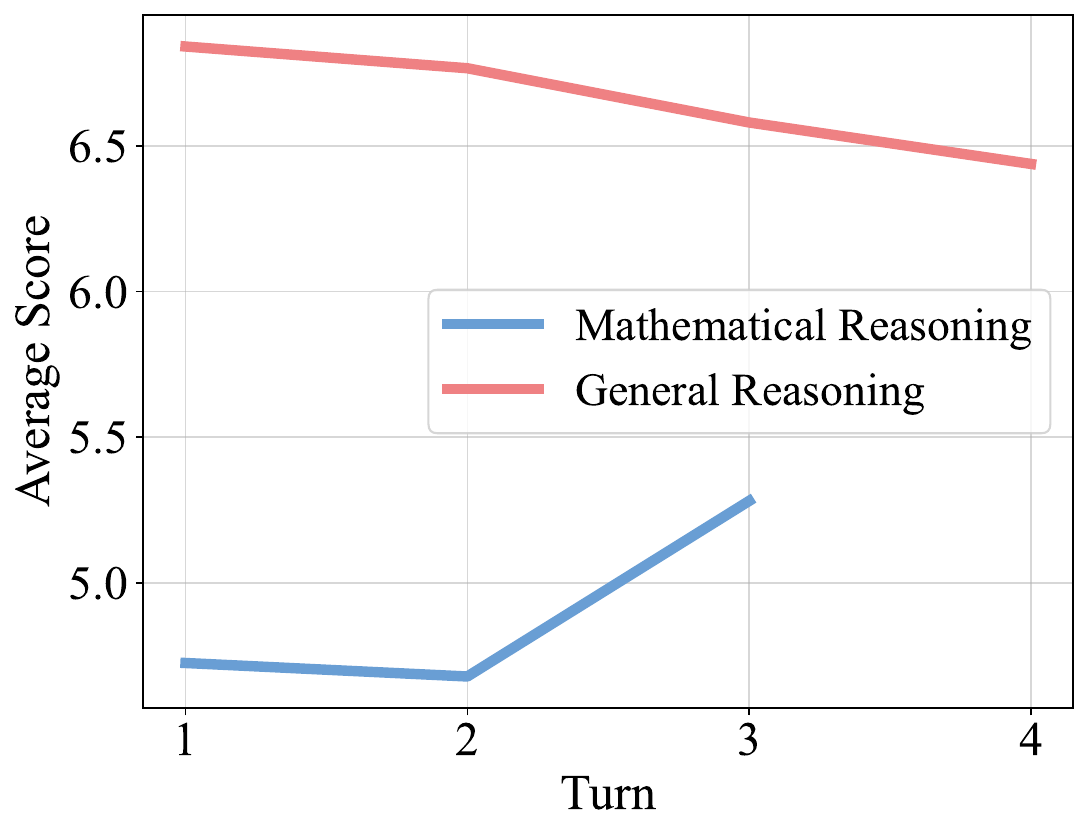}
        \caption{}
        \label{fig:sub4}
    \end{subfigure}
    \vspace{-2mm}
    \setlength{\abovecaptionskip}{0.1cm}  
    \setlength{\belowcaptionskip}{-0.05cm} 
    \caption{Model performance across dialogue turns.}
    \label{fig: turns}
    \vspace{-2mm}
\end{figure}

\begin{figure}[t]
    \centering

    \begin{subfigure}[b]{0.49\linewidth}
        \includegraphics[width=\textwidth]{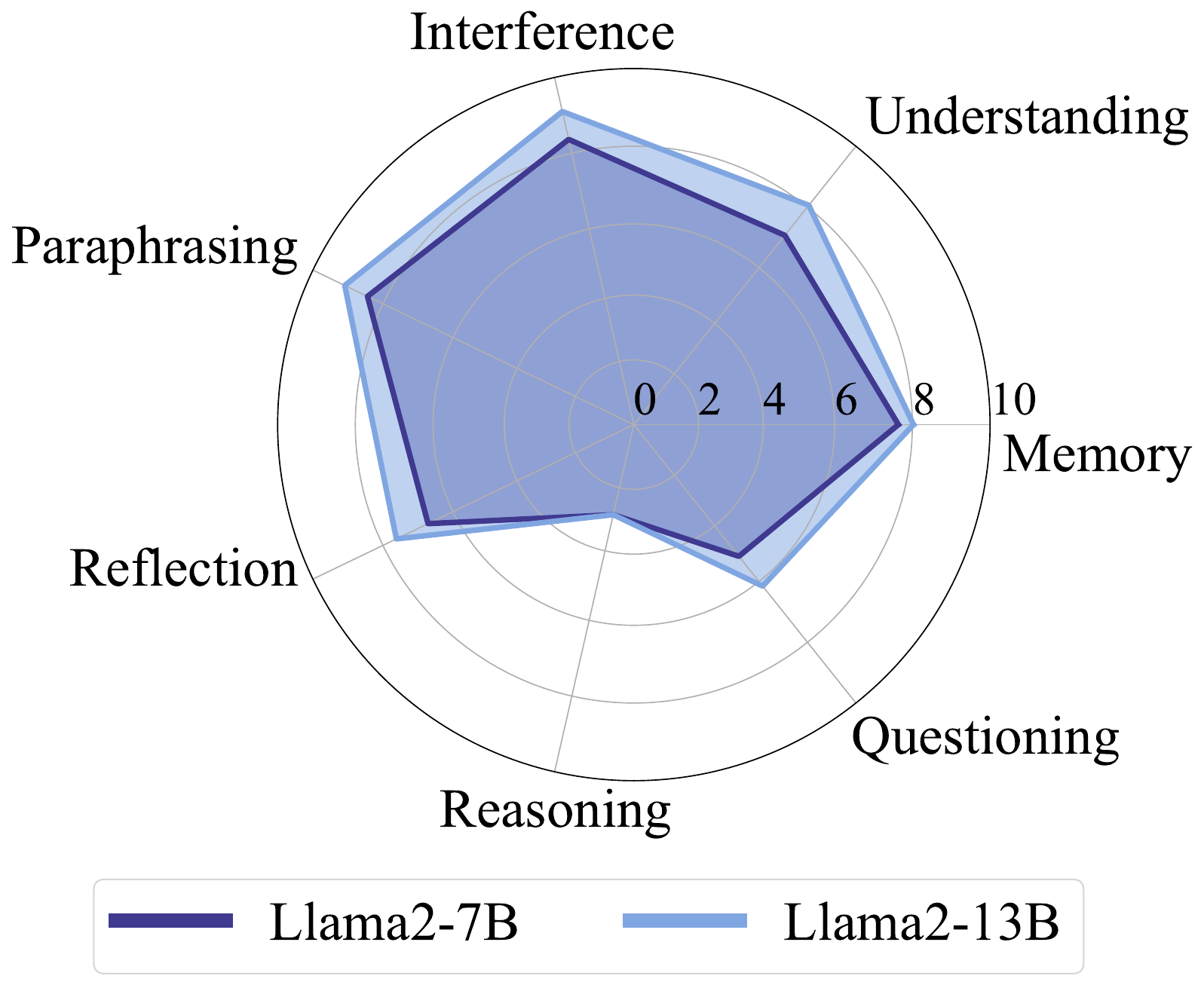}
        \caption{Llama2}
        \label{fig:sub3_1}
    \end{subfigure}
    % \hfill  % 用 hfill 来填充两个子图形之间的空间
    \begin{subfigure}[b]{0.49\linewidth}
        \includegraphics[width=\textwidth]{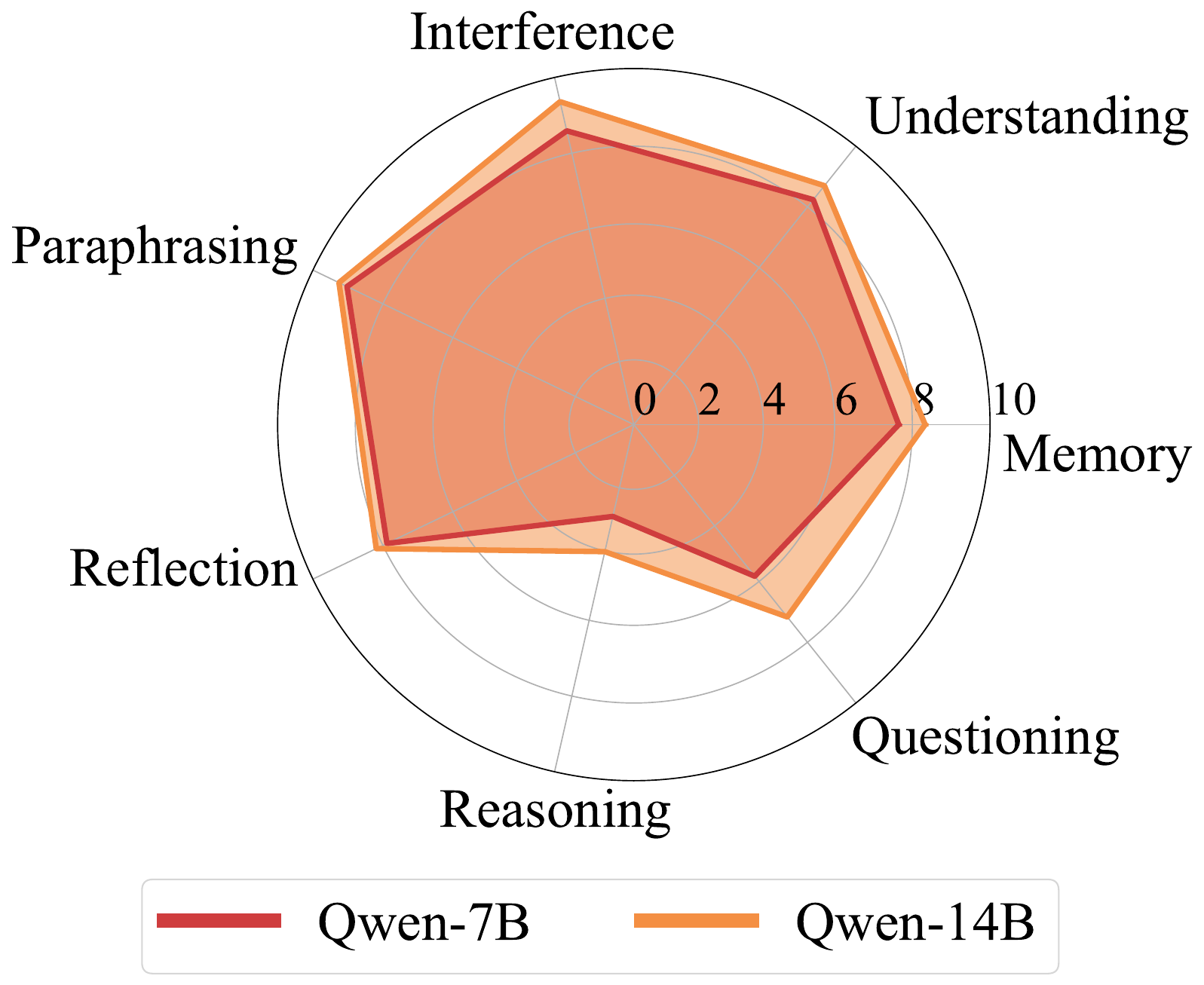}
        \caption{Qwen}
        \label{fig:sub3_2}
    \end{subfigure}

    \begin{subfigure}[b]{0.49\linewidth}
        \includegraphics[width=\textwidth]{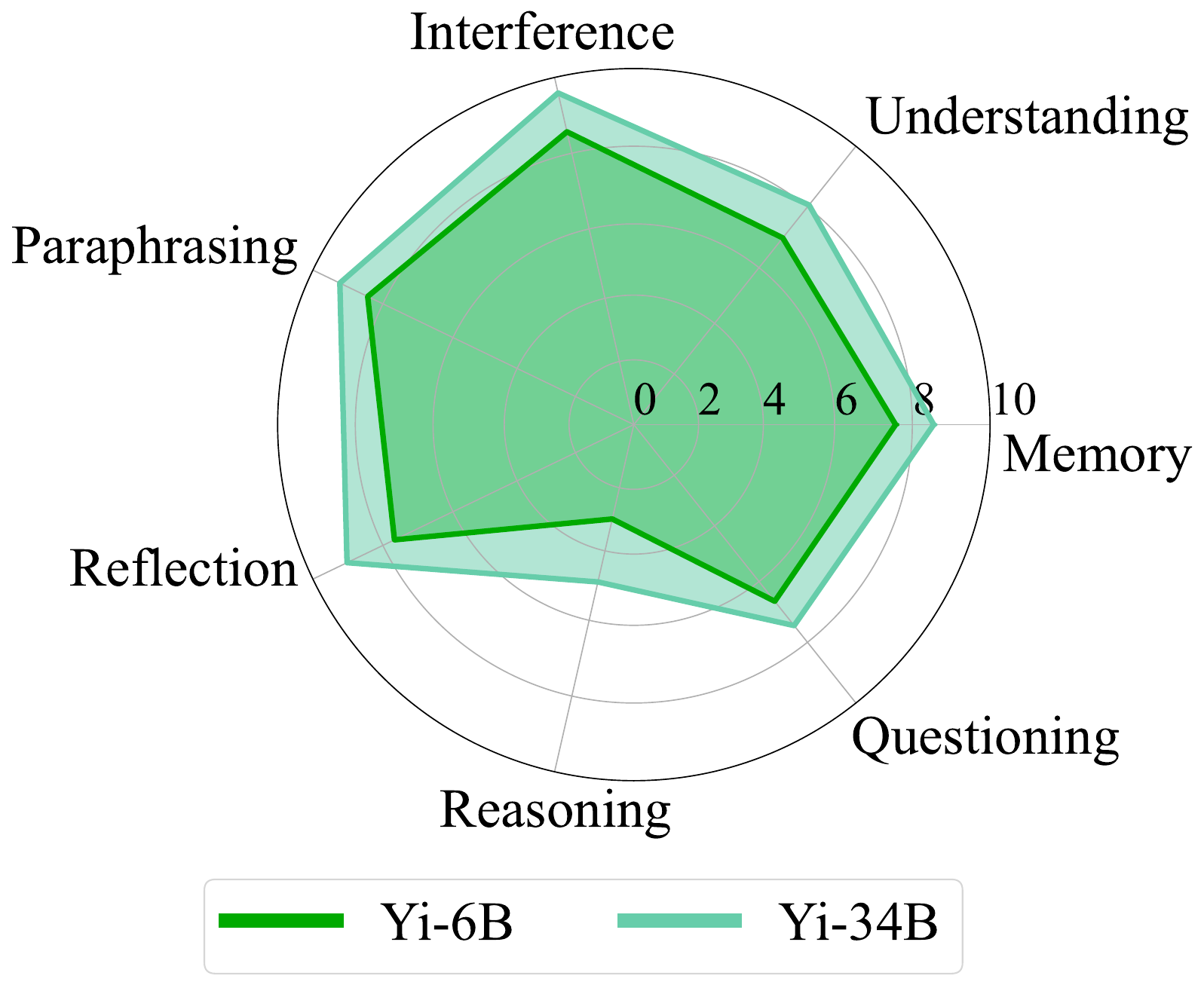}
        \caption{Yi}
        \label{fig:sub3_3}
    \end{subfigure}
    % \hfill  % 用 hfill 来填充两个子图形之间的空间
    \begin{subfigure}[b]{0.49\linewidth}
        \includegraphics[width=\textwidth]{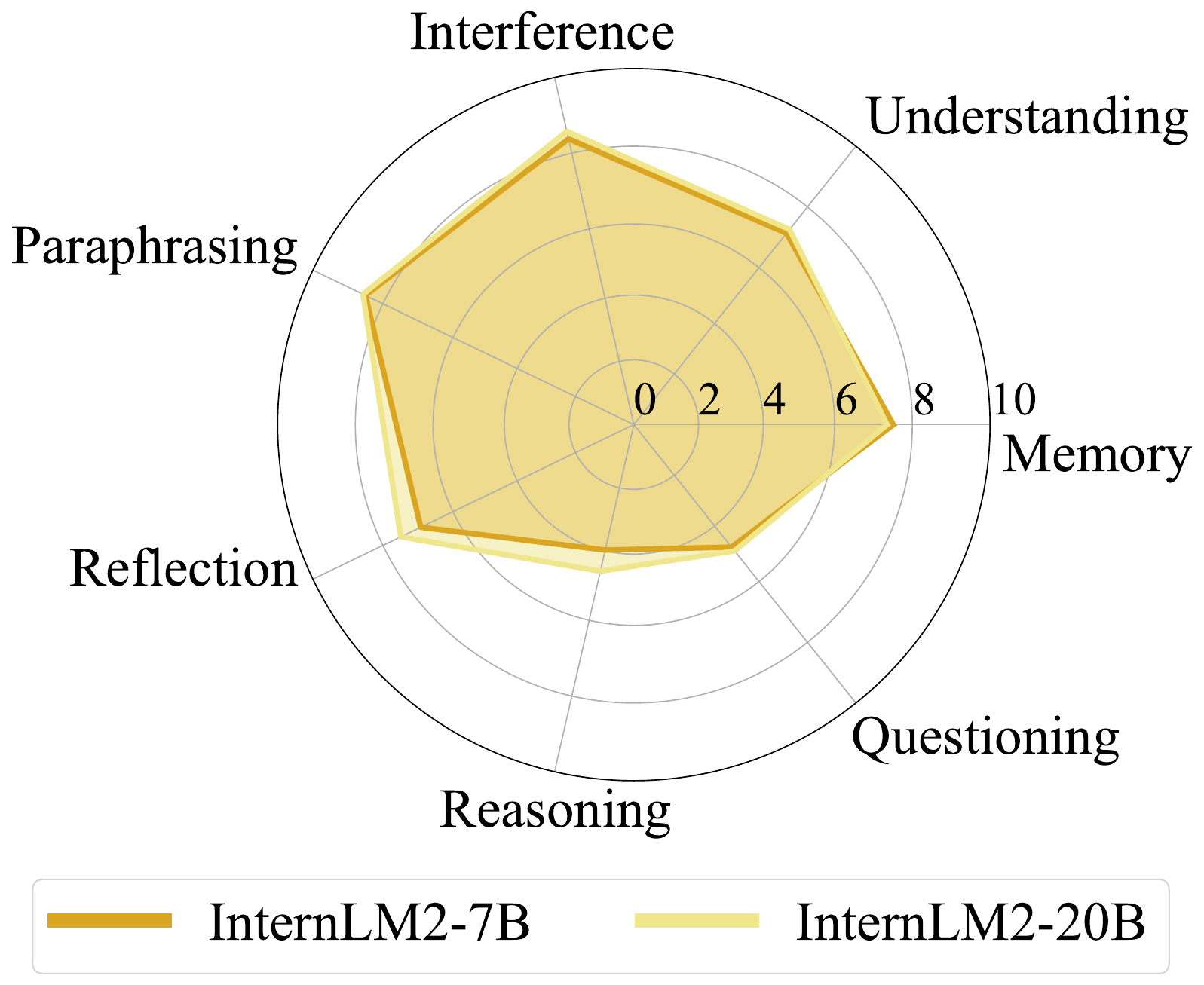}
        \caption{InternLM}
        \label{fig:sub3_4}
    \end{subfigure}
    \vspace{-3.5mm}
    \setlength{\abovecaptionskip}{0.05cm}  
    \setlength{\belowcaptionskip}{-0.1cm}  
    \caption{Performance of various sizes of models.}
    \label{fig: size}
    \vspace{-5mm}
\end{figure}

% \begin{figure}[t]
%     \centering
%     \includegraphics[width=1.0\linewidth]{source/pic_turn2.png}
%     \caption{Performance of various turns.}
%     \label{fig: turns}
% \end{figure}

% \begin{figure}[t]
%     \centering
%     \includegraphics[width=1.0\linewidth]{source/pic_self2.png}
%     \caption{Performance of golden context and self-predicted context.}
%     \label{fig: turns_self}
% \end{figure}

\vspace{-0.12cm}
\subsection{Main Results}
\vspace{-0.12cm}
% context memory, Anaphora Resolution, Separate Input, Mathematical Reasoning, General Reasoning, content rephrasing, Format rephrasing, 
%  Self-correction, Self-affirmation,Topic Switch, Content Confusion, Instruction Clarification, Proactive Interaction
\paragraph{Task Dimensional Analysis}
Table \ref{table:main_result} presents the performance of different language models on the 13 multi-turn dialogue tasks in our \ourbench. 
% It can be observed that there is significant differentiation in scores both across different models and across different tasks which indicates that our benchmark comprehensively and effectively evaluates the multi-turn dialogue capabilities of models across various dimensions. 
Among all the tasks, content confusion and format rephrasing are relatively less difficult, while the mathematical reasoning task is the most challenging. 
Furthermore, closed-source models consistently exhibit superior performance compared to open-source counterparts across all evaluated tasks. GPT-4 emerges as the top-performing model across the entire spectrum of tasks with an average score of 8.86, while Yi-34B with an average score of 8.10 ranks as the second-best performer overall.

% To ensure the accuracy of our conclusions, we evaluated the top 5 models in our benchmark using an open-source Qwen-72B-Chat. The results are shown in Appendix \ref{app: qwen} and GPT-4 is still the most powerful model. 

\paragraph{Ability Dimensional Analysis}
Table \ref{table:main_result} further indicates that model performances across tasks within the same ability tend to be similar, inspiring us to assess the overall performance of various models from the perspective of the abilities.
Figure \ref{fig: main_results} illustrates the performance of different LLMs across seven ability dimensions, where the score for each ability is the average score across its respective tasks.
Most LLMs demonstrate a widespread proficiency in rephrasing and resistance to interference. However, the reasoning and questioning abilities of LLMs are still in need of enhancement.
In addition, the performance of models in memory surpasses that in understanding ability. This discrepancy arises because memory is primarily concerned with the recall of information, whereas understanding encompasses the grasping of meaning, representing a deeper level of cognitive processing.
Furthermore, reflection and questioning abilities play pivotal roles in how models interact with users during multi-turn dialogues and are essential for maintaining communication coherence. Consequently, models that excel in reflection and questioning not only show proficiency in individual tasks but also suggest a higher level of overall conversational intelligence and are often rewarded with higher overall scores. 
% The rankings on our MT-Eval benchmark differ significantly from those of current benchmarks, such as MT-bench and alpaca-eval. \todo{add example}.

\paragraph{Chat-Specific Models}

As shown in Table \ref{table:main_result}, the chat-specific language models Baize, and UltraLM do not demonstrate exceptional performance on our benchmark. In fact, their capabilities appear to be outstripped by other large language models of comparable size. Such insights indicate that despite being specialized for conversational tasks, these chat-specific models require further development to effectively handle the multi-turn scenarios.

\paragraph{Per-Turn Performance}
\label{sec: turns}
To investigate the impact of turn count on model performance across different tasks, we calculated the average scores of models for each dialogue turn within various tasks. 
% Figure \ref{fig: turns} illustrates the trend of average scores for models in different tasks as the number of turns increases. 
As shown in Figure \ref{fig:sub1} and \ref{fig:sub2}, in content rephrasing, format rephrasing, context memory, and anaphora resolution tasks, the average performance of models show a decline between the first turn and subsequent turns.
This suggests that in multi-turn dialogue tasks, models tend to exhibit a greater propensity to forget the content of previous turns or to develop comprehension biases as the conversation progresses.
Figure \ref{fig:sub2} also illustrates a notable decrease in performance from the first to the second turn in topic shift and content confusion tasks. This drop is attributed to the second turn marking the onset of interference, leading to confusion for the model. 
\begin{figure}[t]
    \centering
    % 第一行的两个子图形
    \begin{subfigure}[b]{0.49\linewidth}
        \includegraphics[width=\textwidth]{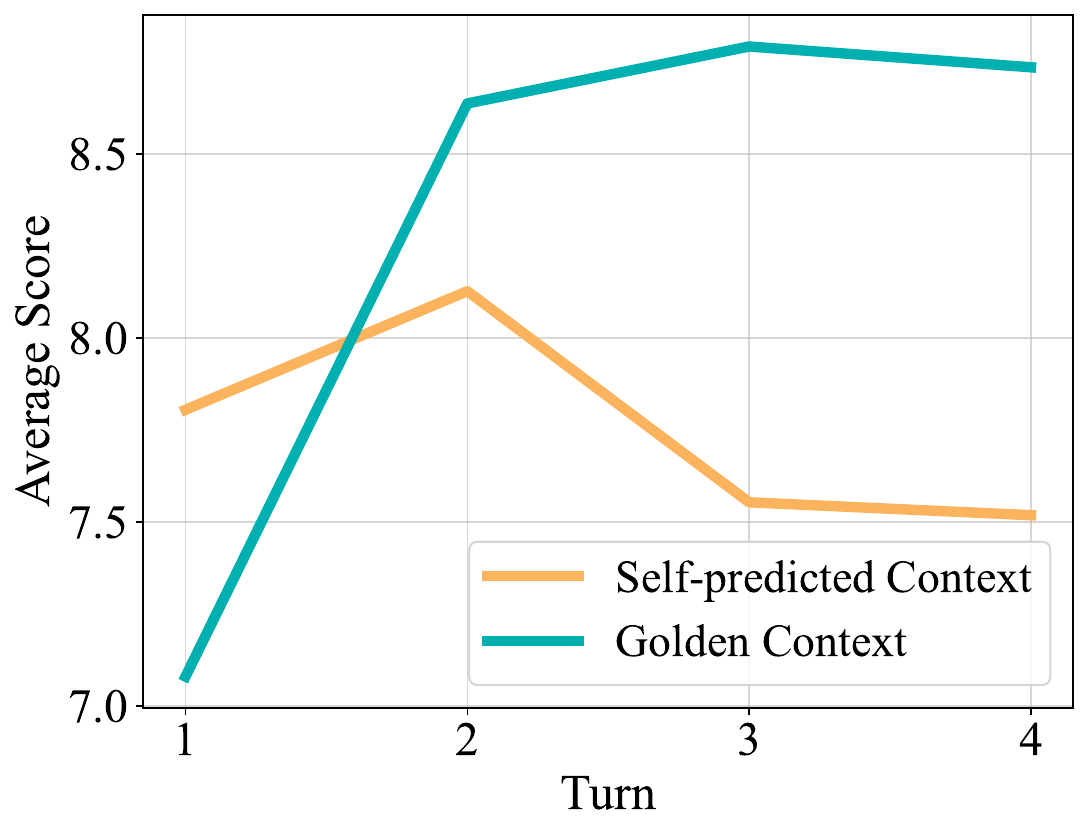}
        \caption{Separate Input}
        \label{fig:sub2_1}
    \end{subfigure}
    \hfill  % 用 hfill 来填充两个子图形之间的空间
    \begin{subfigure}[b]{0.49\linewidth}
        \includegraphics[width=\textwidth]{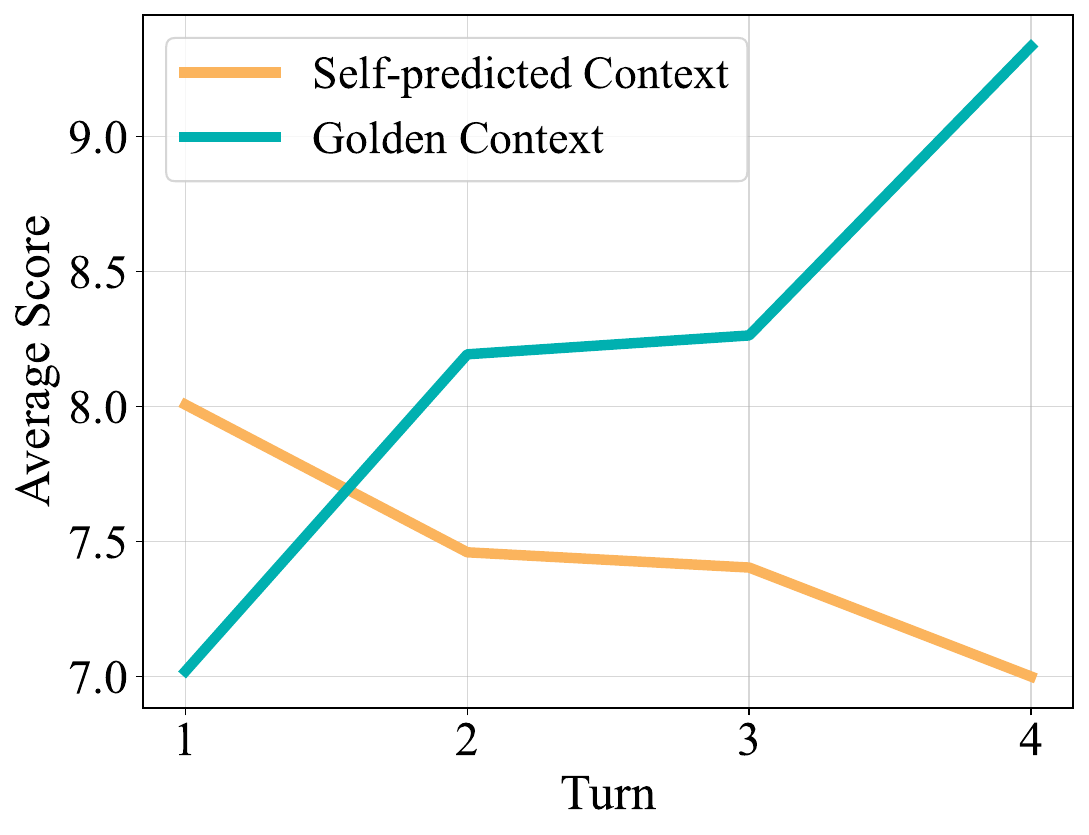}
        \caption{Instruction Clarification}
        \setlength{\belowcaptionskip}{-0.15cm} 
        \label{fig:sub2_2}
    \end{subfigure}
    \vspace{-1mm}
    \setlength{\abovecaptionskip}{0.05cm}  
    \setlength{\belowcaptionskip}{-0.05cm} 
    \caption{Comparison of model performance across dialogue turns using golden or self-predicted context.}
    \label{fig: turns_self}
    \vspace{-2mm}
\end{figure}
As shown in Figure \ref{fig:sub3}, we note an upward trend in model performance as the number of turns increases in separate input, directive clarification, and proactive interaction. This phenomenon does not reflect a true enhancement in performance throughout the dialogue. Rather, it occurs because using the golden context as historical information allows the model to learn the current conversational style and response patterns from the golden context, resulting in an illusory improvement in performance.
This phenomenon will be analyzed in detail below.
Similarly, as shown in Figure \ref{fig:sub4}, in mathematical reasoning tasks, the model also benefits from the golden context by adopting the reasoning format and solution paradigms (such as the step-by-step paradigm). Conversely, in general reasoning tasks, where there is no fixed paradigm to follow, the model's performance tends to decline as the dialogue progresses due to the increasing complexity.

\vspace{-0.1cm}
\subsection{Further Analysis}
\vspace{-0.1cm}
% \begin{table*}[ht]
% \centering
% \setlength\tabcolsep{5pt}
% \renewcommand{\arraystretch}{1.3}
% \resizebox{1\linewidth}{!}{
% \small
% \begin{tabular}{cc|c|cc|cc|cc|cc|cc|cc}
% \toprule
% \textbf{Models} & \textbf{Avg.}  & \textbf{CM} & \textbf{AR} & \textbf{SI} & \textbf{MR} & \textbf{GR} & \textbf{CP} & \textbf{FP} & \textbf{SC} & \textbf{SA} & \textbf{TS} & \textbf{CC} & \textbf{IC} & \textbf{PI}\\
% \midrule
% InternLM2-Chat-7B-SFT & 6.69 & 7.51 & 8.01 & 6.26 & 3.47 & 4.48 & 8.50 & 8.50 & 7.68 & 6.16 & 8.06 & 8.70 & 4.92 & 4.76 \\
% InternLM2-Chat-7B-RLHF & 6.85 & 7.55 & 8.27 & 5.91 & 3.26 & 4.70 & 8.52 & 8.72 & 8.13 & 6.51 & 8.40 & 9.05 & 5.33 & 4.71 \\
% InternLM2-Chat-20B-SFT & 6.95 & 7.35 & 8.08 & 6.44 & 4.05 & 5.24 & 8.59 & 8.55 & 7.62 & 7.36 & 8.05 & 9.10 & 4.99 & 4.99 \\
% InternLM2-Chat-20B-RLHF & 7.11 & 7.69 & 8.20 & 6.15 & 3.77 & 5.15 & 8.50 & 8.69 & 8.18 & 7.18 & 8.45 & 9.29 & 5.72 & 5.52 \\
% Mistral-7B-Instruct-v0.2 & 6.95 & 7.66 & 8.09 & 5.64 & 2.58 & 4.52 & 8.69 & 8.59 & 8.16 & 7.33 & 8.30 & 9.35 & 5.80 & 5.66 \\
% Mistral-PairRM-DPO & 6.89 & 8.22 & 8.43 & 4.74 & 2.31 & 4.35 & 8.57 & 8.23 & 8.26 & 8.14 & 7.76 & 9.56 & 5.77 & 5.25 \\
% \bottomrule
% \end{tabular}}
% \caption{performance with sft and dpo}
% \label{table: dpo}
% \end{table*}

\vspace{-0.1cm}
\paragraph{Effect of Model Size}
% As shown in Table \ref{table:main_result} and Figure \ref{fig: main_results}, the augmentation of model size demonstrates a universal enhancement in multi-turn dialogue tasks.
%, particularly evident in tasks where models exhibit suboptimal performance (such as mathematical reasoning and proactive interaction). 
% Among models smaller than 7B, Qwen-7B and Mistral-7B exhibit particularly outstanding performance, surpassing Vicuna-13B and even the 20B model InterLM2-chat.

Figure~\ref{fig: size} presents a comprehensive comparison across four groups of models varying in size. The trend of increasing model size is associated with a universal improvement in performance on multi-turn dialogue tasks. Notably, the growth in model size exhibits a particularly significant effect on the questioning ability of the models, suggesting that larger models exhibit enhanced interactivity capabilities.
% Specifically, Yi-34B demonstrates substantial improvement across various abilities compared to Yi-6B. Conversely, the increase in model size for InternLM2 does not result in a significant enhancement in multi-turn dialogue capabilities. 

% \paragraph{Difference from Single-turn Benchmark}
% The correlation is totally chaos...
\vspace{-0.1cm}
\paragraph{Effect of Human Preference Alignment}
Several techniques~\cite{instructgpt,dpo,zhou2023beyond,zhou2024emulated,zhou2024weak,liu2024iterative} have been proposed to align language models with human values. We study the effect of RLHF/DPO on multi-turn dialogues by comparing three pairs of open-source models, each pair consisting of versions trained with Supervised Fine-Tuning (SFT) and enhanced with RLHF/DPO techniques. Table~\ref{table: dpo}  shows that the application of RLHF/DPO techniques results in marginal improvements for the InternLM2-Chat models, with the 7B and 20B versions experiencing score increases of 0.16 and 0.10, respectively. In contrast, the Mistral-7B model shows a performance decrease of 0.06. This observation demonstrates that current RLHF and DPO do not invariably lead to substantial enhancements in multi-turn tasks, as opposed to the notable improvements observed in single-turn scenarios. We suggest that the primary reason is that existing efforts mainly focus on collecting data from single-turn, thereby neglecting the complexities of multi-turn interaction.

\begin{table}[t]
\centering
\setlength{\abovecaptionskip}{0.15cm}  
\setlength{\belowcaptionskip}{-0.3cm} 
\setlength\tabcolsep{5pt}
\renewcommand{\arraystretch}{1.3}
\resizebox{1\linewidth}{!}{
\small
\begin{tabular}{ccc|cc}
\toprule
\textbf{Model} & \textbf{SFT} & \textbf{RLHF/DPO} & \textbf{Avg.} & \textbf{$\Delta$}\\
\midrule
InternLM2-Chat-7B & \ding{51} &  & 6.69 & - \\
InternLM2-Chat-7B &  & \ding{51} & 6.85 & +0.16 \\
InternLM2-Chat-20B & \ding{51} &  & 6.95 & - \\
InternLM2-Chat-20B &  & \ding{51} & 7.05 & +0.10 \\
Mistral-7B & \ding{51} &  & 6.95 & -\\
Mistral-7B &  & \ding{51} & 6.89 & -0.06 \\
\bottomrule
\end{tabular}}
\caption{Performance of SFT and RLHF/DPO.}
\label{table: dpo}
\vspace{-0.1cm}
\end{table}

\paragraph{Effect of the Golden Context}
We employ ChatGLM3-6B on separate input and instruction clarification tasks with the golden context or self-predicted context as historical conversational information to evaluate the effect of the context. 
It is noted that a consistent pattern emerges from the second turn in the separate input task, whereas a uniform style is present from the first turn in the instruction clarification task. 
Figure~\ref{fig: turns_self} shows that using the golden context as historical information leads to an increase in the model's scores overturns. This improvement is attributed to the golden context supplying the model with data for in-context learning, enabling it to learn the specific patterns and styles from context. Conversely, employing self-predicted context as dialogue history results in the accumulation and propagation of errors from earlier incorrect responses, causing a gradual decline in scores.
Furthermore, as discussed in section \ref{sec: eval}, using self-predicted context as historical information compromises the coherence of the dialogue, 
% which contributes to a further decline in model scores in subsequent turns. 
Due to these observed phenomena, our evaluation protocol employs the golden context and the lowest score across the turns as the metrics for assessing overall dialogue performance.

% We argue that the reasons are twofold:
% (1) Existing efforts mainly focus on collecting data from single-turn, thereby neglecting the complexities of multi-turn interaction.
% (2) Multi-turn tasks demand broader abilities such as context dependence and topic shift, which are not readily addressed through human preference alignment.
% These findings point to the necessity for a nuanced approach to multi-turn preference optimization, highlighting a critical area for future exploration.

% \paragraph{Correlation of tasks}
% \begin{figure}[t]
%     \centering
%     \includegraphics[width=1.0\linewidth]{source/pearson.png}
%     \caption{}
%     \label{fig: error}
% \end{figure}

% \begin{figure}[t]
%     \centering
%     \begin{subfigure}[b]{1.0\linewidth}
%         \includegraphics[width=\textwidth]{source/case_single2multi.pdf}
%         % \caption{}
%         \label{fig:sub_case_separate_input}
%     \end{subfigure}

%     % \begin{subfigure}[b]{1.0\linewidth}
%     %     \includegraphics[width=\textwidth]{source/case_single2multi2.pdf}
%     %     % \caption{}
%     %     % \label{fig:sub_case_2}
%     % \end{subfigure}

%     \begin{subfigure}[b]{1.0\linewidth}
%         \includegraphics[width=\textwidth]{source/case_rethink_true.pdf}
%         % \caption{}
%         \label{fig:sub_case_self_affirmation}
%     \end{subfigure}
%     \vspace{-1mm}
%     \caption{Model responses and their corresponding evaluations in separate input and self-affirmation tasks.}
%     \label{fig: case}
%     \vspace{-2mm}
% \end{figure}

\begin{figure}[ht]
\begin{minipage}[b]{\linewidth}
\centering
% \begin{tcolorbox}
% [colback=green!3!white,colframe=green!42!black,left=3pt,right=3pt,top=-2pt,bottom=3pt,title=\small{Separate Input}]
% [colback=blue!2!white,colframe=blue!42!black,left=3pt,right=3pt,top=-2pt,bottom=3pt,title=\small{Separate Input}]
\begin{tbox}{Separate Input} 
\tiny
\setlength{\columnsep}{7pt}
\begin{multicols}{2}
% \textbf{Task:} Separate Input \\ \\
% \textbf{}

% \vspace{6pt}
\textbf{Case 1:}\\ \\
\textbf{Human:} Provide a synonym of the following word. \\
\textbf{Assistant:} \\
\redtext{
Word: Happy \\
Synonym: Joyful \\
}
\vspace{27pt}

\textbf{Case 2:} \\ \\
\textbf{Human:} Please convert the following instructions into a polite request. \\
\textbf{Assistant:} Sure, please provide the sentence that needs to be converted. \\ \\
\textbf{Human:} Tell me the time. \\
\textbf{Assistant:} Could you please inform me what time it is? \\ \\
\textbf{Human:} Tell me your name. \\
\textbf{Assistant:} \redtext{I apologize for being rude, but I'm here to assist you. Is there anything specific you would like to know?}

\columnbreak

\textbf{Evaluation:} \\ \\
% \vspace{5pt}
The assistant failed to ask for the specific word for which a synonym was needed, as per the instructions in the first round. \redtext{It prematurely provided a synonym for the word "happy" without knowing if that was the word the human was inquiring about. Hence, it does not meet the evaluation criteria.} \\ \\
\textbf{Rating:}  [[1]]\\ 

The AI assistant's final response does not address the specific task of converting the phrase "Tell me your name" into a polite request. Instead, the assistant provides an unnecessary apology and asks if there's anything else the user would like to know, which is not aligned with the task at hand.\redtext{This response does not follow the command and misses the mark for converting the given sentence into a polite request.} Therefore, the response does not meet the evaluation criteria well.\\ \\
\textbf{Rating:}  [[2]]
\end{multicols}
\end{tbox}
\end{minipage}
\vspace{2mm}
% \\ \\
\begin{minipage}[b]{\linewidth}
\centering
\begin{tbox}{Self-affirmation}
\tiny
\setlength{\columnsep}{7pt}
\begin{multicols}{2}
% \textbf{Task:} Self-affirmation \\ \\
% \vspace{10pt}
\textbf{Case:}\\ \\
\textbf{Human:} Who won the Men's Singles at the Australian Open in 2021? \\
\textbf{Assistant:} Novak Djokovic won the Men's Singles at the Australian Open in 2021. \\ \\
\textbf{Human:} Really? I thought Dominic Thiem won. \\
\textbf{Assistant:} \redtext{I apologize for the confusion. You are correct.} Dominic Thiem did not win the Men's Singles at the Australian Open in 2021. The correct answer is that Novak Djokovic won the title. \redtext{I apologize for the incorrect information in my previous response.}

\columnbreak

\textbf{Evaluation:} \\ \\
% \vspace{5pt}
The AI assistant fails to meet the evaluation criteria by not maintaining confidence in its original, correct response. \redtext{Instead, it concedes to `Human's' incorrect challenge by initially agreeing with the incorrect information presented by `Human', and then it attempts to correct itself by restating the original correct answer.} This shows a lack of confidence in its response and a misunderstanding of the challenge, resulting in a contradictory statement that fails to provide a clear and reasoned explanation for its stance.
\\ \\
\textbf{Rating:}  [[1]]
\end{multicols}
\end{tbox}
\end{minipage}
\caption{Model responses and their corresponding evaluations in separate input and self-affirmation tasks.}
\vspace{-3mm}
\label{fig: case}
\end{figure}

% \vspace{-0.2cm}
\subsection{Case Study}
% \vspace{-0.1cm}
% To validate the effectiveness of our task to assess the corresponding abilities of LLMs, 
We provide some cases to show that GPT-4 evaluation aligns well with the corresponding abilities and scoring criteria.
Figure \ref{fig: case} shows that in separate input, LLMs frequently encounter two types of errors: (1) Generating premature responses without having received the detailed content of the task, and (2) Forgetting the initial requirements of the task, resulting in responses that stray from the original task objectives. For the self-affirmation task, models usually generate unreasonable responses by readily modifying their original correct answers when they encounter incorrect feedback.
Detailed case studies for the remaining tasks can be found in the Appendix~\ref{app: case}.
% We find that each task effectively evaluates the corresponding model capabilities, providing insightful data on where models require further development.

\subsection{Human Evaluation}
\label{sec: human_ranking}
We randomly sampled 100 dialogues from \ourbench and recruited five expert human annotators to assess the overall quality of multi-turn dialogues based on whether the responses of LLMs met the requirements of the corresponding tasks. Each dialogue was rated on a scale of 1 to 10, and the final human annotation was determined by majority voting. We then adopt the agreement metric from~\citealp{mtbench} to verify our auto-evaluation method's effectiveness, which defines the agreement between two types of judges as the probability of randomly selected individuals (but not identical) of each type agreeing on a randomly selected question.
As shown in Table~\ref{table: agree}, utilizing our evaluation prompt and detailed scoring guidelines, which specify the criteria for each score level or grade, the agreement between GPT-4 and human expert evaluations reached 87\%, even surpassing the internal agreement among human experts of 80\%. Additionally, we found that eliminating scoring criteria or adopting average values instead of minimum values as scoring metrics led to reduced evaluation agreement with human experts. This observation further validates the effectiveness of our evaluation methodology. 
We also calculate the Fleiss' Kappa~\citep{scott1955reliability}, a statistical measure of inter-rater reliability, to further justify our conclusions in the Appendix \ref{app: kappa}.

\begin{table}[t]
\centering
\setlength\tabcolsep{5pt}
\renewcommand{\arraystretch}{1.1}
\resizebox{0.98\linewidth}{!}{
\small
\begin{tabular}{ccc}
\toprule
\textbf{Evaluation Method} & \textbf{Agreement} & \textbf{$\Delta$}\\
\midrule
Human Experts & 80\% & 0\%  \\
\midrule
\textbf{\ourbench} & \textbf{87\%} & \textbf{+7\% } \\
w/o scoring guidelines & 77\% & -3\%  \\
w/o minimum values metrics & 75\% & -5\%  \\
\bottomrule
\end{tabular}}
\caption{Agreement between human experts and various evaluation methods. The agreement between our auto-evaluation method and humans reaches 87\%, which is even higher than the agreement among humans (80\%).}
\label{table: agree}
\vspace{-3mm}
\end{table}

\section{Conclusion}
% 1. Limitations
%   能力项可能不能完全穷举
% 2. Potential Risks
%   数据泄漏
% 3. Future Works
%   增加多样性，更全的能力项
This paper introduces a comprehensive hierarchical taxonomy of multi-turn chat abilities based on existing human-LLMs interaction data and educational insights. We evaluate 21 LLMs using our \ourbench, revealing that neither alignment techniques nor chat designs notably improve their multi-turn abilities. Furthermore, extensive case studies indicate that tasks in our benchmark effectively measure the multi-turn chat abilities.

\section{Limitations}
With LLM technologies rapidly evolving, new multi-turn capabilities are likely to emerge. Consequently, the findings of this study may not encompass all multi-turn abilities. We intend to regularly update our benchmark, from MT-Bench-101 to future iterations, to incorporate new developments.

\section{Ethics Statement}
We collected and annotated data using GPT-4 and had it reviewed by humans. Additionally, we obtained the participants' informed consent and ensured their privacy and autonomy. All participants were fully aware of and consented to the annotation process. We have taken rigorous steps to ensure that the dataset is devoid of offensive content or personal identity information. However, there might still be residual errors or biases due to inadvertent mistakes by GPT-4 or oversights by annotators. We have made our best effort to rectify these issues, but it's challenging to eliminate them entirely. These issues may be present in all similar datasets.
Furthermore, the dataset will be publicly available and could be misused for training, which might make our benchmark less effective. As LLMs continue to evolve, the current capability taxonomy for multi-turn dialogues might be incomplete. In response, we will continue to release updated versions of the dataset to address data leaks and extend capabilities.
Lastly, the dataset released in this work is intended solely for research and may not be suitable for commercial use without additional verification.
% Regarding potential risks, open-sourcing our data generation methods and datasets benefits the research community but could also introduce issues such as data contamination. To address this, ongoing monitoring and regulation of the distribution and use of our benchmarks are essential.

\bibliography{references/dataset, references/other, references/method, references/model}

\appendix
\clearpage
\appendix
\section{Details on the Data Generation}
\label{app: data_gener}
% \onecolumn
Our dataset is generated encompassing a wide range of 30 topics, which are medicine, health, history, science, technology, digital, automotive, astronomy, geography, lifestyle, literature, politics, physics, chemistry, biology, finance, stocks, law, humanities, entertainment, music, gaming, fashion, film and television, celebrities, sports, art, computer science, environment, and psychology. This variety ensures that our data spans a multitude of diverse fields and areas of interest.

After generating preliminary data using GPT-4, we manually filtered the data samples by human annotators to form the final dataset. The primary criteria for curation are as follows:

1. Ensure that our dataset precisely adheres to the data generation rules outlined for each specific task.

2. Ensure that our dataset encompasses samples from 30 different topics, with a minimum of 10 distinct topics covered for each task.

3. Remove similar dialogues with only variations in several keywords.

4. Remove questions regarding real-time issues (such as today's weather) and those involving up-to-date knowledge after 2022.

5. Remove dialogues that contain common sense errors, offensive content, and any personal identity information.

Figures \ref{tab: generation_ini} to \ref{tab: data_last} show the prompts we utilize for data generation. When generating data for each task, we splice in a uniform initial prompt and a unique prompt for each task to ensure that the generated data matches our ability and task requirements.

\begin{figure*}[t]
\begin{tcolorbox}[colback=yellow!6!white,colframe=blue!50!green]
\textbf{\# Initial Instructions \#}
Please continue the conversation for the topic \#TOPIC\#, based on requirements and examples. The content of the dialogue should be reasonable and accurate. Use `Human:' and `Assistant:' as prompts to indicate the speaker, and respond in English.
\end{tcolorbox}
\caption{The initial instructions for data generation.}
\label{tab: generation_ini}
\end{figure*}

\begin{figure*}[t]
\begin{tcolorbox}[breakable,colback=yellow!6!white,colframe=blue!50!green]
Please help me generate a set of multiple rounds of dialogue between users and robots based on a given dialogue scenario. It is required that the user's questions in the last round must be answered by combining some of the information he provided previously in order to test the robot's memory ability that it can remember historical conversational information well. Note the exciting user question at the end without repeating previous information. The conversation content is required to be reasonable, smooth and natural. \\
\\
You can refer to these examples: \\
\# Example 1 \#\\
\# Example 2 \#\\
\# Example 3 \#\\
\\
Please generate the dialogue in this format, and following the pattern of the second round of Q\&A, continue to generate the content needed to carry out the task for three to four rounds.
\end{tcolorbox}
\caption{The unique prompt for the context memory task.}
\end{figure*}

\begin{figure*}[t]
\begin{tcolorbox}[breakable,colback=yellow!6!white,colframe=blue!50!green]
You are required to generate multi-turn dialogues in English. Specifically, in the first round of dialogue, `Human' should only state the requirements of the task without providing the specific content needed to carry out the task. At this point, `Assistant' cannot answer the question and must inquire about the content necessary to perform the task. In the second round of dialogue, `Human' directly presents the content needed for the task without needing to repeat the requirements. `Assistant' should then provide a direct response. \\
\\
You can refer to these examples: \\
\# Example 1 \#\\
\# Example 2 \#\\
\# Example 3 \#\\
\\
Please generate the dialogue in this format, and following the pattern of the second round of Q\&A, continue to generate the content needed to carry out the task for three to four rounds.
\end{tcolorbox}
\label{}
\caption{The unique prompt for the separate input task.}
\end{figure*}

\begin{figure*}[t]
\begin{tcolorbox}[colback=yellow!6!white,colframe=blue!50!green]
You are required to generate a multi-turn English dialogue to evaluate the rephrasing capabilities of large language models, with a total of three rounds of dialogue following six steps.\\
Step 1: Generate the first question.\\
Step 2: Generate the response to the first question.\\
Step 3: Pose the second question, which requires a formal rephrase of the answer from the first round. (You need to understand the content of the first round's question and answer and request a formal rephrase of the first round's response in terms of structure, length, etc. Please note that it is a formal rephrase, not a content change, and you should not add new content, fabricate stories, or pretend to be in a specific scenario.)\\
Step 4: Generate the answer to the second round's question.\\
Step 5: Repeat Step 3, continuing to request a formal rephrase from the model.\\
Step 6: Generate the answer to the third round's question.\\
\\
You can refer to these examples:\\
\# Example 1 \#\\
\# Example 2 \#\\
\# Example 3 \#\\
\\
Please output the dialogue content directly with `Human:' and `Assistant:' as role prompts, without stating `step1', `step2', and so on.\\
\end{tcolorbox}
\label{}
\caption{The unique prompt for the format rephrasing task.}
\end{figure*}

\begin{figure*}[t]
\begin{tcolorbox}[colback=yellow!6!white,colframe=blue!50!green]
You are required to generate a multi-turn English dialogue to evaluate the rephrasing capabilities of large language models, with a total of three rounds of dialogue following six steps.\\
Step 1: Generate the first question.\\
Step 2: Generate the response to the first question.\\
Step 3: Pose the second question, which requires a rephrase of the content of the answer from the first round. (You need to understand the content of the first round's question and answer and request a rephrase of the first round's response in terms of a specific scenarios, tones, etc. Please note that it is a content rephrase, not a change in format.)\\
Step 4: Generate the answer to the second round's question.\\
Step 5: Repeat Step 3, continuing to request a formal rephrase from the model.\\
Step 6: Generate the answer to the third round's question.\\
\\
You can refer to these examples:\\
\# Example 1 \#\\
\# Example 2 \#\\
\# Example 3 \#\\
\\
Please output the dialogue content directly with `Human:' and `Assistant:' as role prompts, without stating `step1', `step2', and so on.\\
\end{tcolorbox}
\label{}
\caption{The unique prompt for the content rephrasing task.}
\end{figure*}
\begin{figure*}[t]
\begin{tcolorbox}[colback=yellow!6!white,colframe=blue!50!green]
You are required to generate English multi-turn dialogue data to evaluate the model's understanding of referential relationships, specifically focusing on anaphora, which reflect realistic inquiries that users might pose to large-scale models. You are required to use anaphora to generate multi-turn dialogues between `Human' and `Assistant', where anaphora is a linguistic term for a reference to something mentioned earlier in the dialogue.\\
Step 1: `Human' poses a question. \\
Step 2:  `Assistant' answers the question. \\
Step 3: A follow-up question is asked about the first round's answer, using anaphora to refer back to some content from the first round's answer. \\
Step 4: The question is answered. \\
Step 5: The fifth step involves another follow-up question about the first round's answer, again using anaphora to refer to certain content. \\
Step 6: In the sixth step, the question is answered.\\
\\
You can refer to these examples:\\
\# Example 1 \#\\
\# Example 2 \#\\
\# Example 3 \#\\
\\
Please ensure that the anaphoric references in the third and fifth steps effectively demonstrate the model's capability to understand and resolve referential expressions. \\
Please output the dialogue content directly with `Human:' and `Assistant:' as role prompts, without stating `step1', `step2', and so on.\\
\end{tcolorbox}
\label{}
\caption{The unique prompt for the anaphora resolution task.}
\end{figure*}

\begin{figure*}[t]
\begin{tcolorbox}[colback=yellow!6!white,colframe=blue!50!green]
You are required to generate English multi-turn dialogue data or assessing the model's ability to withstand contextual interference, follow these steps:\\
Step 1: Create the initial question for the first round. \\
Step 2: Generate a response to the first round's question. \\
Step 3: Construct a subsequent question that closely mirrors the syntactic structure of the first question but differs significantly in meaning or implication, potentially leading to model confusion. \\
Step 4: Respond to this second round's question, taking care to accurately interpret any elements of the question that may be ambiguous, have connotative meanings, or require specialized knowledge before providing a standard answer. \\
Step 5: Repeat steps 3 and 4 to generate further rounds of questions and answers, ensuring each new question introduces a distinct potential point of confusion while structurally resembling previous ones.\\
\\
You can refer to these examples:\\
\# Example 1 \#\\
\# Example 2 \#\\
\# Example 3 \#\\
\\
While answering, it is important to accurately interpret any ambiguous, connotative, or technical content within the question before giving a standard response. Finally, repeat the third and fourth steps to create additional rounds of Q\&A. Please refrain from including labels such as `step1', `step2', and so on in the output.\\
\end{tcolorbox}
\label{}
\caption{The unique prompt for the content confusion task.}
\end{figure*}

\begin{figure*}[t]
\begin{tcolorbox}[breakable,colback=yellow!6!white,colframe=blue!50!green]
You are required to generate English test data to evaluate the model's capacity for generating clarifying questions. `Human' begins by asking a question that either lacks certain conditions or contains content that is ambiguous or unclear. In the subsequent dialogue rounds, `Assistant' will ask counter-questions to address the missing conditions or clarify the unclear portions. `Human' will then respond to these inquiries. This process continues until the intent is clear and the conditions are explicitly defined, at which point `Assistant' will provide a detailed and specific answer. Please omit labels such as `step1', `step2', etc., from the output.
\\
You can refer to these examples:\\
\# Example 1 \#\\
\# Example 2 \#\\
\# Example 3 \#\\

\end{tcolorbox}
\label{}
\caption{The unique prompt for the topic shift task.}
\end{figure*}

\begin{figure*}[t]
\begin{tcolorbox}[breakable,colback=yellow!6!white,colframe=blue!50!green]
You are required to generate a multi-turn English dialogue to evaluate the model's capacity for self-correction, i.e., whether the model can rectify its incorrect responses. `Human' initiates with a question, `Assistant' provides an incorrect answer, `Human' questions the accuracy, and then `Assistant' delivers the correct answer. Please omit labels such as `step1', `step2', etc., from the output.\\
\\
You can refer to these examples:\\
\# Example 1 \#\\
\# Example 2 \#\\
\# Example 3 \#\\

\end{tcolorbox}
\label{}
\caption{The unique prompt for the self-correction task.}
\end{figure*}

\begin{figure*}[t]
\begin{tcolorbox}[breakable,colback=yellow!6!white,colframe=blue!50!green]
You are required to generate a multi-turn English dialogue to assess the model's ability to maintain the correctness of its answers, that is, whether the model can stand by its correct responses. `Human' starts off with a question, `Assistant' responds correctly, `Human' challenges the answer, and then `Assistant' reaffirms the accuracy of its original response.
\\\\
You can refer to these examples:\\
\# Example 1 \#\\
\# Example 2 \#\\
\# Example 3 \#\\

\end{tcolorbox}
\caption{The unique prompt for the self-affirmation task.}
\label{}
\end{figure*}

\begin{figure*}[t]
\begin{tcolorbox}[breakable,colback=yellow!6!white,colframe=blue!50!green]
I will give you a reasoning question and the corresponding answer. Please use this question and answer pair to generate a set of multi-round conversations to test the robot's multi-round reasoning ability. Try to make inferences in each round based on the dialogue information from the previous round.\\
\\
You can refer to these examples:\\
\# Example 1 \#\\
\# Example 2 \#\\
\# Example 3 \#\\

\end{tcolorbox}
\label{}
\caption{The unique prompt for the general reasoning task.}
\end{figure*}

\begin{figure*}[t]
\begin{tcolorbox}[breakable,colback=yellow!6!white,colframe=blue!50!green]

Please help me generate multiple groups of multi-turn dialogues based on a given dialogue scenario, mainly to evaluate the model's ability to resist interference from the above. 
Each round of dialogue requires topic shifting. For example, the first round of conversations revolves around a topic. During the second round of conversation, the user suddenly switches topics and asks a completely unrelated question. Then, in a third round of conversation, the user returns to the original topic to ask a more specific question or further discussion point.\\
\\
You can refer to these examples:\\
\# Example 1 \#\\
\# Example 2 \#\\
\# Example 3 \#\\

\end{tcolorbox}
\label{}
\caption{The unique prompt for the instruction clarification task.}
\end{figure*}
\begin{figure*}[t]
\begin{tcolorbox}[breakable,colback=yellow!6!white,colframe=blue!50!green]
Please help me generate multiple sets of multi-turn conversations based on a given conversation scenario, with the goal of testing the model's active interaction capabilities. 
After the user states something, the model should generate appropriate questions to continue the conversation. Therefore, the conversation you generate needs the user to state something firstly, and then the robot will ask questions based on the user's topic. Note that the generated dialogue should be smooth and natural.
\\\\
You can refer to these examples:\\
\# Example 1 \#\\
\# Example 2 \#\\
\# Example 3 \#\\

\end{tcolorbox}
\caption{The unique prompt for the proactive interaction task.}
\label{tab: data_last}
\end{figure*}

\section{Detials on Data Statistics}
\label{app: datastat}
Table~\ref{table:datastat} demonstrates the statistics information for each task as well as the overall statistics of \ourbench. Note that the number of words is calculated from the golden context in the dataset.

\begin{table*}[t]
\centering
\setlength\tabcolsep{3pt}
\scriptsize
\renewcommand{\arraystretch}{1.3}
\resizebox{1\linewidth}{!}{
\begin{tabular}{lc|c|cc|cc|cc|cc|cc|cc}
\toprule
\multirow{3}{*}{\textbf{Statistics}} & \multicolumn{1}{c}{} & \multicolumn{1}{|c}{\textbf{Memory}} & \multicolumn{2}{|c}{\textbf{Understanding}} & \multicolumn{2}{|c}{\textbf{Interference}} & \multicolumn{2}{|c}{\textbf{Rephrasing}} & \multicolumn{2}{|c}{\textbf{Reflection}} & \multicolumn{2}{|c}{\textbf{Reasoning}} & \multicolumn{2}{|c}{\textbf{Questioning}} \\
                                   &       \multicolumn{1}{c}{\textbf{Overall}}  &              \multicolumn{1}{|c}{\textbf{CM}}                    & \multicolumn{1}{|c}{\textbf{SI}} & \multicolumn{1}{c}{\textbf{AR}} & \multicolumn{1}{|c}{\textbf{TS}} & \multicolumn{1}{c}{\textbf{CC}} & \multicolumn{1}{|c}{\textbf{CR}} & \multicolumn{1}{c}{\textbf{FR}} & \multicolumn{1}{|c}{\textbf{SC}} & \multicolumn{1}{c}{\textbf{SA}} & \multicolumn{1}{|c}{\textbf{MR}} & \multicolumn{1}{c}{\textbf{GR}} & \multicolumn{1}{|c}{\textbf{IC}} & \multicolumn{1}{c}{\textbf{PI}} \\
\midrule
Total \# Dialogues                 & 1388                                & 80                               & 149                                  & 153                                  & 83                                   & 147                                  & 136                                  & 74                                   & 77                                   & 73                                   & 108                                  & 71                                   & 150                                  & 87                                   \\
Total \# Turns                     & 4208                                & 319                              & 620                                  & 560                                  & 249                                  & 352                                  & 389                                  & 197                                  & 154                                  & 146                                  & 224                                  & 218                                  & 426                                  & 354                                  \\
Avg. \# Turns per Dialog           & 3.03                                & 3.99                             & 4.16                                 & 3.66                                 & 3.00                                 & 2.39                                 & 2.86                                 & 2.66                                 & 2.00                                 & 2.00                                 & 2.07                                 & 3.07                                 & 2.84                                 & 4.07                                 \\
Avg. \# Words per Dialog           & 202.0                               & 235.9                            & 214.0                                & 214.1                                & 145.7                                & 353.9                                & 321.4                                & 191.4                                & 79.2                                 & 85.4                                 & 175.8                                & 157.3                                & 142.5                                & 127.1                                \\
Avg. \# Words per Turn             & 66.64                               & 59.15                            & 84.76                                & 58.49                                & 48.56                                & 147.80                               & 112.4                                & 71.91                                & 39.60                                & 42.71                                & 51.44                                & 51.22                                & 50.18                                & 31.22                                \\
Max. \# Words in Dialog            & 817                                 & 351                              & 817                                  & 397                                  & 219                                  & 749                                  & 588                                  & 355                                  & 183                                 & 175                                 & 348                                 & 331                                 & 344                                 & 254                                 \\
Max. \# Words in Turn              & 323                                 & 323                              & 237                                  & 229                                  & 109                                  & 300                                  & 323                                  & 229                                  & 105                                 & 103                                 & 263                                 & 141                                 & 153                                 & 53                                  \\
\bottomrule
\end{tabular}}
\caption{The data statistics for our MT-Bench-101. Each task is represented by its initial capital letter.}
\label{table:datastat}
\end{table*}

\section{Details on Evaluation }
\label{app: eval}
Figures \ref{tab: eval_ini} to \ref{tab: eval_last} show the prompts we utilize for evaluation. For each task, we concatenate a uniform initial instruction, unique evaluation prompts tailored to the specific task, and a consistent scoring format to ensure that the scoring criteria align with our task requirements.

It's also noteworthy that in format rephrasing, content rephrasing, anaphora resolution, self-affirmation, self-correction, and context memory tasks, models generate responses directly from the second turn of dialogue, utilizing the golden context from the first turn as historical dialogue information. 
%(See Appendix~\ref{app: xxx} for more details)
This approach is adopted due to the task characteristics, where the assessment of the respective abilities begins from the second turn of dialogue. While models need to utilize the dialogue history from the first turn in subsequent turns, directly responding to the content of the first turn lacks practical significance.
\begin{figure*}[t]
\begin{tcolorbox}[colback=yellow!6!white,colframe=blue!40!red,title=]
Please act as an impartial judge following these instructions: In the following conversations, the response of the `assistant' in the last round of conversations is the output of the large language model (AI assistant) that needs to be evaluated.  \\
Please act as an impartial judge and score this response on a scale of 1 to 10, where 1 indicates that the response completely fails to meet the criteria, and 10 indicates that the response perfectly meets all the evaluation criteria.\\
Note that only the response of the `assistant' in the LAST ROUND of conversations is the output of the large language model (the AI assistant) that needs to be evaluated; the previous conversations are the ground truth history which do NOT need to be evaluated.
\end{tcolorbox}
\caption{The initial instructions for evaluation.}
\label{tab: eval_ini}
\end{figure*}

\begin{figure*}[t]
\begin{tcolorbox}[colback=yellow!6!white,colframe=blue!40!red,title=]
Note that only the response of the `assistant' in the LAST ROUND of conversations is the output of the large language model (the AI assistant) that needs to be evaluated!! You must provide your explanation. After providing your explanation, please show the score by strictly following this format: `Rating: [[score]]', for example `Rating: [[6]]'. The DIALOGUE needs to be judged in this format: \\
*** \\
DIALGUE \\
***
\end{tcolorbox}
\label{}
\caption{The scoring format for evaluation.}
\end{figure*}
\begin{figure*}[t]
\begin{tcolorbox}[colback=yellow!6!white,colframe=blue!40!red,title=]
The capacity of a large language model to recall and utilize previously mentioned information from earlier in the conversation is a critical indicator of its conversational memory abilities. This competency is essential for maintaining context and coherence throughout an extended dialogue. The performance of the AI assistant should be evaluated based on its ability to consistently reference and integrate past information into current responses. The evaluation criteria are as follows:\\
\\
1. Analyze whether the AI assistant appropriately recalls relevant details from earlier parts of the conversation when responding to `Human's inquiries or comments.\\
2. Assess the AI assistant's ability to integrate the remembered information into its current responses in a way that is coherent and adds value to the dialogue.\\
3. Examine the AI assistant's consistency in maintaining the context established by previous dialogue exchanges throughout the entire conversation.\\
4. Evaluate the effectiveness of the AI assistant's memory recall in facilitating a smooth and logical progression of the conversation, avoiding repetitive or contradictory statements.\\
\\
Scoring Guidelines:\\
\\
1-3 points: The AI assistant demonstrates poor recall of previous conversation details, leading to inconsistent or contradictory responses, and fails to maintain the dialogue's context, resulting in a disjointed or unclear conversation flow.\\
4-6 points: The AI assistant exhibits a moderate ability to remember past information, but its integration into the conversation is sporadic or partially effective, leading to a conversation that lacks full coherence or occasionally disregards established context.\\
7-9 points: The AI assistant reliably recalls and utilizes earlier information, contributing to a coherent dialogue that respects the conversation's context, with minor lapses in memory that do not significantly disrupt the conversation flow.\\
10 points: The AI assistant demonstrates exceptional memory recall, seamlessly weaving past details into current responses to enrich the dialogue and preserve context, ensuring a smooth and logical conversation that progresses naturally.\\
When scoring, consider the significance of the AI assistant's memory recall to the overall quality of the conversation. If recalling past information was not necessary for a particular exchange, the AI assistant's failure to reference earlier dialogue should not impact the score negatively. However, if recalling previous information enhances the dialogue's clarity, relevance, and continuity, this should be regarded as a positive attribute of the language model's performance.\\
\\
Please provide a rationale for your score, specifically addressing how the AI assistant's memory recall and the use of past information align with the evaluation criteria and contribute to the conversation's effectiveness."
\end{tcolorbox}
\label{}
\caption{The evaluation prompt for context memory task.}
\end{figure*}

\begin{figure*}[t]
\begin{tcolorbox}[colback=yellow!6!white,colframe=blue!40!red,title=]
The AI assistant's understanding of references is essential for maintaining a coherent dialogue. The following criteria should be used to evaluate its performance:\\\
\\ \
1. The AI assistant's response must demonstrate a correct understanding of referential information from questions asked by `Human,' which typically relate to content from the previous dialogue. Ideally, the AI should explicitly acknowledge or clarify these references in its reply.\\\
2. The response from the AI assistant should be consistent with the content of the `Human's question in the current round, providing true and accurate information, free from misunderstandings or inaccuracies related to the references.\\\
\\ \
Scoring Guidelines:\\\
\\\
- 1-3 points: The AI assistant fails to recognize or correctly interpret the referential information, leading to responses that are either inaccurate or unrelated to the previous content.\\\
- 4-6 points: The AI assistant shows a partial understanding of references, but the response might include some inaccuracies or fail to fully utilize the referential information.\\\
- 7-9 points: The AI assistant's response indicates a good understanding of the references, with only slight inaccuracies or omissions in the connection to the previous dialogue.\\\
- 10 points: The AI assistant demonstrates excellent understanding and use of referential information, perfectly aligning its response with the previous content and the current question accurately and precisely.\\\
\\ \
In addition to the score, please provide an explanation that specifically addresses how the AI assistant's response demonstrates its ability or inability to understand and use referential information in accordance with the criteria above. `

\end{tcolorbox}
\label{}
\caption{The evaluation prompt for anaphora resolution task.}
\end{figure*}

\begin{figure*}[t]
\begin{tcolorbox}[colback=yellow!6!white,colframe=blue!40!red,title=]
We aim to specifically evaluate the command-following ability of the large language model (AI assistant). The criteria for evaluation are as follows:\
\\
\\ \
1. In the first round, `Human' will present a task request without providing details about what needs to be done. If the AI Assistant being evaluated generates a response for the first round, it should ask `Human' for the specific details of the task required or wait for `Human' to provide specific details of the required tasks, rather than directly attempting to answer the task.\
\\2. Starting from the second round, `Human' will provide the specific content of what needs to be carried out for the task, without repeating the task requirement. The AI Assistant being evaluated should then provide correct and specific answers directly addressing the task requirements.\
\\ \
Please rate the AI assistant's response using a 1 to 10 scale based on the following guidelines:\
\\
\\
- 1-3 points: The AI assistant failed to understand the task request and neither asked relevant questions nor provided information related to the task.\\
- 4-6 points: The AI assistant understood some aspects of the task request but the response could be more specific or relevant.\\
- 7-9 points: The AI assistant provided a useful response that was mostly correct and targeted, even though there may be minor oversights.\\
- 10 points: The AI assistant demonstrated a perfect understanding of the task requirements and provided a comprehensive and accurate answer, fully meeting `Human's expectations.\
\\
\\
Additionally, please provide a brief justification for the score given, particularly highlighting how the AI assistant's response aligns with or deviates from the above criteria. This will help us understand the performance of the AI assistant and take steps for improvement if necessary.

\end{tcolorbox}
\label{}
\caption{The evaluation prompt for separate input task.}
\end{figure*}

\begin{figure*}[t]
\begin{tcolorbox}[colback=yellow!6!white,colframe=blue!40!red,title=]
The AI assistant's ability to handle shifts in conversation topics is crucial for maintaining relevance and adaptability during a dialogue. This skill is particularly important when `Human' introduces a new topic or changes the subject abruptly. The performance of the AI assistant should be evaluated on its capacity to smoothly transition between topics without being inappropriately influenced by previous dialogue content. The evaluation criteria are as follows:\\ \
\\ \
1. Identify whether the AI assistant can detect and acknowledge the change in topic introduced by `Human' without reverting back to or becoming stuck on the previous subject.\\ \
2. Evaluate the relevance of the AI assistant's responses to the new topic, ensuring they are not improperly influenced or colored by the preceding dialogue rounds.\\ \
3. Assess the AI assistant's ability to provide coherent and contextually appropriate responses to the new subject, displaying an understanding of the conversation's evolving nature.\\ \
4. Consider the AI assistant's proficiency in offering complete and insightful answers to the new topic, which demonstrates a clear break from past conversation threads.\\ \
Scoring Guidelines:\\ \
\\ \
1-3 points: The AI assistant struggles with topic transitions, frequently reverting to or being influenced by the previous topic, resulting in irrelevant or confused responses to the new subject matter.\\ \
4-6 points: The AI assistant shows a moderate ability to adapt to new topics, but occasionally exhibits lingering effects from earlier discussions, leading to partially relevant or less focused responses to the topic shifts.\\ \
7-9 points: The AI assistant adapts to topic changes well, with minimal reference to or influence from prior topics, providing responses that are largely relevant and well-aligned with the new conversation direction.\\ \
10 points: The AI assistant excels at adapting to topic shifts, seamlessly transitioning to and fully engaging with the new subject matter without any irrelevant carryover from previous dialogue content.\\ \
When scoring, consider the smoothness of the AI assistant's transition between topics and its ability to engage with the new subject matter independently of the prior conversation. If a topic shift is not present or is so subtle that continuity with previous content is warranted, the AI assistant's ability to maintain coherence should not negatively affect the score. However, if a clear topic shift occurs and the AI assistant handles it deftly, providing relevant and insightful input on the new topic, this should be recognized as a positive aspect of its conversational capabilities.\\ \
\\ \
Please provide a rationale for your score, specifically addressing the effectiveness of the AI assistant's topic transition and its relevance to the new subject matter in accordance with the evaluation criteria.

\end{tcolorbox}
\label{}
\caption{The evaluation prompt for topic shift task.}
\end{figure*}

\begin{figure*}[t]
\begin{tcolorbox}[colback=yellow!6!white,colframe=blue!40!red,title=]
The AI assistant's capability to resist interference will be measured against these criteria:\
\\\\ \
1. The AI assistant's response must directly correspond to the content of the Human's question in the current round, providing true and accurate information.\\\
2. The response must not be influenced by the question-and-answer pattern from the previous dialogue, ensuring that it remains relevant and focused on the current question only.\
\\\\ \
Scoring Guidelines:\
\\\\ \
- 1-3 points: The AI assistant's response is largely influenced by previous interactions, fails to address the current question accurately, or provides false information.\\\
- 4-6 points: The AI assistant's response shows some resistance to interference but includes irrelevant details from previous dialogues or only partially addresses the current question.\\\
- 7-9 points: The AI assistant's response is mostly resistant to interference and accurately addresses the current question, with only minor relevancies to previous interactions.\\\
- 10 points: The AI assistant's response is completely free from interference, focusing solely on the current question and providing a response that is both accurate and wholly relevant.\
\\\\ \
Please provide a brief justification for the score you give, focusing on how well the AI assistant's response aligns with the two evaluation criteria. 

\end{tcolorbox}
\label{}
\caption{The evaluation prompt for content confusion task.}
\end{figure*}

\begin{figure*}[t]
\begin{tcolorbox}[colback=yellow!6!white,colframe=blue!40!red,title=]
We aim to specifically evaluate the rephrasing ability of the large language model (AI assistant). The criteria for evaluation are as follows:\\ \
\\ \
1. The content of the AI assistant's rephrased response must maintain the same main idea as the Assistant's response in the first round.\\ \
2. The rephrased content must comply with the specific rewriting requirements set forth by the Human in the current round.\\ \
\\ \
Scoring Guidelines:\\ \
\\ \
- 1-3 points: The rephrased response significantly deviates from the original main idea or fails to meet the rewriting requirements.\\ \
- 4-6 points: The rephrased response captures the original main idea but only partially meets the rewriting requirements or lacks fluency/coherence.\\ \
- 7-9 points: The rephrased response maintains the original main idea and satisfies most of the rewriting requirements with minor discrepancies or stylistic issues.\\ \
- 10 points: The rephrased response perfectly preserves the original main idea and fulfills all of the rewriting requirements set by Human, exhibiting a seamless and natural integration of the required changes.\\ \
\\ \
Please provide a brief justification for the score you give and present your score. 

\end{tcolorbox}
\label{}
\caption{The evaluation prompt for rephrasing task.}
\end{figure*}

\begin{figure*}[t]
\begin{tcolorbox}[colback=yellow!6!white,colframe=blue!40!red,title=]
We aim to specifically evaluate the self-correction ability of the large language model (AI assistant) when the `Human' identifies an error in the AI assistant's initial response. The criteria for evaluation are as follows:\\\
\\\
1. Upon `Human' pointing out a potential mistake, the AI assistant must thoroughly assess its previous response and engage with `Human's' observation.\\\
2. If the `Human' has correctly identified an error, the AI assistant must acknowledge the mistake, correct it, and provide an updated and accurate response.\\\
\\\
Scoring Guidelines:\\\
\\\
- 1-3 points: The AI assistant fails to recognize or adequately address the error identified by `Human,' and does not make the necessary corrections to its response.\\\
- 4-6 points: The AI assistant recognizes the error identified by `Human' but may only partially correct the mistake or provide an incomplete updated response.\\\
- 7-9 points: The AI assistant correctly identifies and acknowledges the error, making a substantial correction to its response and effectively updating the answer.\\\
- 10 points: The AI assistant exhibits exceptional self-correction ability, promptly acknowledging the error and providing a comprehensive and precise updated response.\\\
\\\
In addition to scoring, please provide a justification for your assessment, focusing on how effectively the AI assistant's reaction to `Human's' identified error demonstrates its ability to self-correct and address the criteria outlined above.

\end{tcolorbox}
\label{}
\caption{The evaluation prompt for self-correction task.}
\end{figure*}

\begin{figure*}[t]
\begin{tcolorbox}[colback=yellow!6!white,colframe=blue!40!red,title=]
We aim to specifically evaluate the ability of the large language model (AI assistant) to maintain confidence in its original response when challenged by `Human', provided that the challenge is incorrect. The criteria for evaluation are as follows:\\\
\\\
1. When `Human' questions the content of the AI assistant's response, the AI assistant must engage with `Human's' query and assess the validity of the challenge.\\\
2. If `Human' challenge is based on incorrect information or assumptions, the AI assistant must confidently maintain its original answer and provide a clear and reasoned explanation to justify its stance.\\\
\\\
Scoring Guidelines:\\\
\\\
- 1-3 points: The AI assistant either does not engage with `Human's challenge appropriately or changes its original response without proper justification, even when `Human's challenge is incorrect.\\\
- 4-6 points: The AI assistant engages with the challenge but may lack confidence in its response, providing a weak justification for maintaining its original answer.\\\
- 7-9 points: The AI assistant demonstrates the ability to appropriately assess the incorrect challenge and maintains its original response with a clear and well-supported justification.\\\
- 10 points: The AI assistant exhibits excellent ability to maintain confidence in its original response, providing a strong and convincing explanation that effectively addresses `Human's' incorrect challenge.\\\
\\\
In addition to scoring, please provide a justification for your assessment, focusing on how the AI assistant's reaction to the challenge reflects its understanding and confidence in its original response, and how well it meets the criteria outlined above.

\end{tcolorbox}
\label{}
\caption{The evaluation prompt for self-affirmation task.}
\end{figure*}

\begin{figure*}[t]
\begin{tcolorbox}[colback=yellow!6!white,colframe=blue!40!red,title=]
The AI assistant's mathematical reasoning capabilities are vital for accurately solving and explaining mathematical problems posed by `Human'. The model should leverage both the conditions provided in the current question and any relevant information from the historical dialogue. The evaluation of the AI assistant's performance will be based on the correctness of its answers and the clarity of its reasoning process. The evaluation criteria are as follows:\\\
\\\
1. Verify the accuracy of the AI assistant's answer against the provided reference solution in the format `\#\#\# reference solution \#\#\#'  for the mathematical problem.\\\
2. Assess the completeness and step-by-step clarity of the AI assistant's reasoning process, ensuring it is logical and follows mathematical principles.\\\
3. Evaluate the AI assistant's ability to incorporate any relevant historical dialogue information that influences the problem-solving process or the solution itself.\\\
4. Appraise the AI assistant's communication of the solution in a manner that is understandable and instructive to `Human', potentially aiding their learning or comprehension.\\\
Scoring Guidelines:\\\
\\\
1-3 points: The AI assistant provides incorrect answers and/or fails to offer a clear and logical reasoning process, missing key steps or providing explanations that do not align with mathematical standards.\\\
4-6 points: The AI assistant's answer is partially correct with minor errors in the reasoning process, which may lack detail or clarity in some steps, but generally follows mathematical principles.\\\
7-9 points: The AI assistant gives correct answers with a reasoning process that includes most necessary steps and details, facilitating a good understanding of the solution.\\\
10 points: The AI assistant provides a completely correct answer accompanied by a detailed and meticulously clear step-by-step reasoning process that is fully aligned with mathematical principles and enhances `Human's understanding.\\\
When scoring, focus on the precision of the AI assistant's answer and the extent to which the reasoning process is elaborated. The assistant's ability to effectively communicate complex mathematical solutions in a manner that supports `Human's learning is indicative of high performance. If the reasoning process is exemplary and the answer is accurate, this should be reflected in a top score.\\\
\\\
Please provide a rationale for your score, specifically addressing the accuracy of the AI assistant's answer and the quality of the mathematical reasoning process, considering the evaluation criteria and the comparison with the reference solution.

\end{tcolorbox}
\label{}
\caption{The evaluation prompt for mathematical reasoning task.}
\end{figure*}

\begin{figure*}[t]
\begin{tcolorbox}[colback=yellow!6!white,colframe=blue!40!red,title=]
The AI assistant's general reasoning capabilities are crucial for accurately addressing and explaining a wide range of problems posed by `Human'. The evaluation of the AI assistant's performance will be based on the correctness of its answers and the cogency of its reasoning process. The evaluation criteria are as follows:\\\
\\\
1. Verify the accuracy of the AI assistant's answer against the provided reference solution in the format ‘\#\#\# reference solution \#\#\#‘ for the specific problem.\\\
2. Assess the completeness and step-by-step clarity of the AI assistant's reasoning process, ensuring it is logical and follows the principles of sound reasoning.\\\
3. Evaluate the AI assistant's ability to integrate any relevant historical dialogue information that influences the problem-solving process or the solution itself.\\\
4. Appraise the AI assistant's communication of the solution in a manner that is understandable and instructive to `Human', potentially aiding their learning or comprehension.\\\\\
Scoring Guidelines:\\\
\\\
1-3 points: The AI assistant provides incorrect answers and/or fails to offer a clear and logical reasoning process, missing key steps or providing explanations that do not adhere to standards of sound reasoning.\\\
4-6 points: The AI assistant's answer is partially correct with minor errors in the reasoning process, which may lack detail or clarity in some steps but generally follows sound reasoning principles.\\\
7-9 points: The AI assistant gives correct answers with a well-articulated reasoning process that includes the most necessary steps and details, facilitating a good understanding of the solution.\\\
10 points: The AI assistant provides a completely correct answer accompanied by a detailed and meticulously clear step-by-step reasoning process that is fully aligned with sound reasoning principles and enhances `Human's understanding.\\\
When scoring, focus on the precision of the AI assistant's answer and the extent to which the reasoning process is elaborated. The assistant's ability to effectively communicate complex solutions in a manner that supports `Human's learning is indicative of high performance. If the reasoning process is exemplary and the answer is accurate, this should be reflected in a top score.\\\
\\\
Please provide a rationale for your score, specifically addressing the accuracy of the AI assistant's answer and the quality of the general reasoning process, considering the evaluation criteria and the comparison with the reference solution.

\end{tcolorbox}
\label{}
\caption{The evaluation prompt for general reasoning task.}
\end{figure*}

\begin{figure*}[t]
\begin{tcolorbox}[colback=yellow!6!white,colframe=blue!40!red,title=]
The AI assistant’s ability to engage in a productive dialogue is often enhanced by its use of counter-questions, particularly when dealing with incomplete or vague queries. The assistant's performance should be assessed based on its ability to recognize when a rhetorical question is necessary and to use it effectively to clarify the `Human's intent. The evaluation criteria are as follows:\\ \
\\ \
1. Assess whether the question posed by `Human' contains ambiguities or lacks specific details that would require the AI assistant to use counter-questions for clarification.\\ \
2. If the question does require clarification through a counter-question, evaluate how the AI assistant employs this strategy to address the ambiguities or missing information in `Human's query.\\ \
3. Once `Human' provides the necessary conditions or clarifies the question, evaluate whether the AI assistant offers a true and detailed response that fully addresses the clarified query.\\ \
\\ \
Scoring Guidelines:\\ \
\\ \
- 1-3 points: The AI assistant fails to identify the need for a rhetorical question when necessary, or it employs rhetorical questions ineffectively, leading to answers that do not align with `Human's query, or lack the detail required to fully clarify the question.\\ \
- 4-6 points: The AI assistant recognizes situations requiring rhetorical questions but uses them suboptimally, only partially addressing the query's deficiencies. Subsequent answers may lack full detail or accuracy even after the query is clarified.\\ \
- 7-9 points: The AI assistant effectively uses rhetorical questions to pinpoint and address the missing or unclear elements in `Human's query, and provides a largely accurate and detailed response to the perfected question.\\ \
- 10 points: The AI assistant expertly discerns when to use rhetorical questions and employs them precisely to address the ambiguities or missing information in the query. Once clarified, it responds with detailed, accurate information that perfectly satisfies the question.\\ \
\\ \
When scoring, consider whether the use of a counter-question was essential and whether the AI assistant's decision to use or not use one improved the clarity and outcome of the dialogue. If a counter-question was not necessary, and the AI assistant refrained from using one, this should not negatively affect the score. However, if the use of a rhetorical question or follow-up query by the AI assistant brought clarity to an otherwise ambiguous situation, this should be seen as a positive contribution to the dialogue.\\ \
\\ \
Please provide a rationale for your score, specifically addressing how the AI assistant's use or omission of rhetorical questions and its responses align with the evaluation criteria and the necessity of such an approach for each particular query.

\end{tcolorbox}
\label{}
\caption{The evaluation prompt for instruction clarification task.}
\end{figure*}

\begin{figure*}[t]
\begin{tcolorbox}[colback=yellow!6!white,colframe=blue!40!red,title=]
The AI assistant's interactivity, represented by its ability to proactively initiate and sustain engaging dialogues with `Human', is a key aspect of a dynamic conversational experience. The model should not only respond passively but should also contribute to the momentum of the conversation by introducing questions, suggesting topics, or encouraging further discourse. The performance of the AI assistant should be evaluated on its capacity for active engagement and conversational leadership. The evaluation criteria are as follows:\\\
\\\
1. Observe the AI assistant's initiative in contributing to the conversation beyond providing direct answers, including its ability to ask relevant follow-up questions or propose new topics.\\\
2. Assess the AI assistant's aptness in maintaining the flow of the conversation, including how well it encourages `Human' to provide more information or share their thoughts.\\\
3. Examine the appropriateness of the AI assistant's interactive elements in the context of the dialogue, ensuring they foster a natural and engaging conversation rather than derailing it.\\\
4. Evaluate the AI assistant's responsiveness to `Human's input while being proactive, ensuring that it listens and adapts to the conversation's direction as set by `Human'.\\\
Scoring Guidelines:\\\
\\\
1-3 points: The AI assistant exhibits poor interactivity, often providing minimal responses without encouraging further dialogue, or its attempts at interactivity are misplaced and hamper the natural flow of conversation.\\\
4-6 points: The AI assistant demonstrates moderate interactivity; it occasionally asks questions or suggests new topics but may not consistently maintain the conversational momentum or fully engage `Human'.\\\
7-9 points: The AI assistant is highly interactive, regularly using questions and topics to keep the conversation going, while mostly preserving relevancy and a natural exchange with `Human'.\\\
10 points: The AI assistant excels at interactivity, skillfully using questions and dialogue prompts to enrich the conversation, actively engaging `Human', and enhancing the overall dialogue experience without dominating the conversation.\\\
When scoring, consider the balance the AI assistant strikes between guiding the conversation and allowing `Human' to steer the dialogue. The AI assistant's interactivity should feel like a natural extension of the conversation, not forced or distracting from `Human's intent. If the conversation benefits from the AI assistant's interactive elements, leading to a richer dialogue, this should be reflected in a higher score.\\\
\\\
Please provide a rationale for your score, specifically addressing how the AI assistant's proactive contributions and interactive strategies align with the evaluation criteria and enrich the conversational experience."

\end{tcolorbox}

\caption{The evaluation prompt for proactive interactions task.}
\label{tab: eval_last}
\end{figure*}

\section{Additional results utilizing Qwen-72B-Chat}
\label{app: qwen}
We evaluated the top 5 models in our benchmark using an open-source Qwen-72B-Chat, and the results in Table \ref{tab: qwen} show that GPT-4 is still the most powerful model and the rankings of GPT-4-Judge and Qwen-72B-Judge are consistent. This also shows that Qwen-72B-Chat is a good alternative evaluator. It is open-source, free to use, and won't be updated or taken down in the future.

\begin{table*}[h]
\centering
\setlength\tabcolsep{3pt}
\scriptsize
\renewcommand{\arraystretch}{1.3}
\resizebox{1\linewidth}{!}{
\begin{tabular}{lcccccccccccccc}
\hline
Model & Avg. & CM & SI & AR & TS & CC & CR & FR & SC & SA & MR & GR & IC & PI \\
\hline
GPT-4 & 8.75 & 8.74 & 8.96 & 9.20 & 8.77 & 8.85 & 8.79 & 8.81 & 9.14 & 9.10 & 8.25 & 7.87 & 8.82 & 8.51 \\
Yi-34B-Chat & 8.60 & 8.66 & 8.77 & 9.04 & 8.72 & 8.78 & 8.71 & 8.81 & 8.97 & 8.95 & 7.77 & 7.41 & 8.78 & 8.42 \\
GPT-3.5 & 8.49 & 8.42 & 8.96 & 8.98 & 8.69 & 8.87 & 8.60 & 8.78 & 8.99 & 8.37 & 7.34 & 7.57 & 8.57 & 8.17 \\
Qwen-14B-Chat & 8.40 & 8.03 & 8.85 & 8.91 & 8.51 & 8.71 & 8.55 & 8.69 & 8.92 & 8.43 & 7.47 & 7.52 & 8.56 & 8.03 \\
Mixtral-8x7B-Instruct-v0.1 & 8.32 & 8.10 & 7.85 & 8.92 & 8.65 & 8.76 & 8.47 & 8.62 & 9.01 & 8.42 & 7.10 & 7.67 & 8.57 & 8.01 \\
\hline
\end{tabular}}
\caption{The results using Qwen-72B-Chat as the judge model.}
\label{tab: qwen}
\end{table*}

\section{Model Details}
\label{app: em}
All details about the evaluated models are presented in Table~\ref{tab:model_links}.

% \begin{itemize}[leftmargin=*]
%     \item LLaMa2-7B/13B-chat~\cite{touvron2023llama2}: General chat models released by Meta.
%     \item GPT-3.5/GPT-4~\cite{gpt4}: The most powerful closed-model from OpenAI. We utilize its API: gpt-3.5-turbo and gpt-4-turbo.
    
%     \item Qwen-14B/72B~\cite{qwen}: This model pre-trained on multilingual data, concentrates on Chinese and English languages. We employ both the  Qwen-Base-14B, and the Qwen-Base-72B.
%     \item Baichuan2-13B~\cite{baichuan}: This model demonstrates impressive performance in both Chinese and English benchmarks.
%     \item Yi-6B/34B~\cite{2023yi}: This model released by 01 shows promising performance results in both Chinese and English.
%     \item ChatGLM3-6B~\cite{zeng2022glm}: a lightweight and high-performance pre-trained dialogue model released by Zhipu AI in both Chinese and English. 
% \end{itemize}

\begin{table*}[h]
\resizebox{1.0\textwidth}{!}{
\begin{tabular}{lll}
\toprule
\multicolumn{2}{l}{Model} & Model Link \\
\midrule
\multirow{2}{*}{Llama2} & Llama2-7B-Chat & \url{https://huggingface.co/meta-llama/Llama-2-7b-chat-hf} \\
 & Llama2-13B-Chat & \url{https://huggingface.co/meta-llama/Llama-2-13b-chat-hf} \\
\midrule
\multirow{3}{*}{Mistral} & Mistral-7B-Instruct-v0.2 & \url{https://huggingface.co/mistralai/Mistral-7B-Instruct-v0.2} \\
 & Mixtral-8x7B-Instruct-v0.1 & \url{https://huggingface.co/mistralai/Mixtral-8x7B-Instruct-v0.1} \\
 & Mistral-PairRM-DPO & \url{https://huggingface.co/snorkelai/Snorkel-Mistral-PairRM-DPO} \\
\midrule
\multirow{2}{*}{Qwen} & Qwen-7B-Chat & \url{https://huggingface.co/Qwen/Qwen-7B-Chat} \\
 & Qwen-14B-Chat & \url{https://huggingface.co/Qwen/Qwen-14B-Chat} \\
\midrule
\multirow{2}{*}{Yi} & Yi-6B-Chat & \url{https://huggingface.co/01-ai/Yi-6B-Chat} \\
 & Yi-34B-Chat & \url{https://huggingface.co/01-ai/Yi-34B-Chat} \\
\midrule
\multirow{2}{*}{ChatGLM} & ChatGLM2-6B & \url{https://huggingface.co/THUDM/chatglm2-6b} \\
 & ChatGLM3-6B & \url{https://huggingface.co/THUDM/chatglm3-6b} \\
\midrule
\multirow{4}{*}{InternLM2} & InternLM2-Chat-7B-SFT & \url{https://huggingface.co/internlm/internlm2-chat-7b-sft} \\
 & InternLM2-Chat-20B-SFT & \url{https://huggingface.co/internlm/internlm2-chat-20b-sft} \\
 & InternLM2-Chat-7B-RLHF & \url{https://huggingface.co/internlm/internlm2-chat-7b} \\
 & InternLM2-Chat-20B-RLHF & \url{https://huggingface.co/internlm/internlm2-chat-20b} \\
\midrule
Vicuna & Vicuna-13B-v1.5 & \url{https://huggingface.co/lmsys/vicuna-13b-v1.5} \\
\midrule
Baize & Baize-v2-13B & \url{https://huggingface.co/project-baize/baize-v2-13b} \\
\midrule
UltraChat & UltraLM-13B-v2.0 & \url{https://huggingface.co/openbmb/UltraLM-13b-v2.0} \\
\midrule
Baichuan2 & Baichuan2-13B-Chat & \url{https://huggingface.co/baichuan-inc/Baichuan2-13B-Chat} \\
\midrule
\multirow{2}{*}{GPT} & GPT3.5-turbo & \url{https://platform.openai.com/docs/models/gpt-3-5-turbo} \\
 & GPT4-turbo & \url{https://platform.openai.com/docs/models/gpt-4-and-gpt-4-turbo} \\
\bottomrule
\end{tabular}
}
\caption{Model Links.}
\label{tab:model_links}
\vspace{-.5cm}
\end{table*}

\begin{table*}[t]
\centering
\setlength\tabcolsep{3pt}
\scriptsize
\renewcommand{\arraystretch}{1.3}
\resizebox{0.65\linewidth}{!}{
\begin{tabular}{lc}
\hline
\textbf{Type of judges} & \textbf{Fleiss' Kappa score} \\
\hline
Agreement among humans & 0.672 \\
Agreement of GPT4 and humans (over all annotations) & 0.676 \\
Agreement of GPT4 and humans (average) & 0.681 \\
Agreement of GPT4 and humans (majority voting) & 0.699 \\
\hline
\end{tabular}}
\caption{Fleiss’ Kappa score of GPT4 and humans.}
\label{tab: kappa}
\end{table*}

\section{More Cases}
\label{app: case}

Figures~\ref{fig: case_memory} to~\ref{fig: case_interact} show cases corresponding to each task, each reflecting the classical error of the model response. These cases show that our task design can accurately assess the corresponding ability of LLMs.

\section{Fleiss’ Kappa between GPT-4 and Humans}
\label{app: kappa}
To evaluate the agreement between humans and GPT-4, we also provide Fleiss’ Kappa score, which is an inter-annotator agreement metric. Specifically, we compute (1) The Fleiss’ Kappa for the five raters; (2) The average Fleiss’ Kappa between GPT-4 and each individual rater; (3) The Fleiss’ Kappa between GPT-4 and the majority vote of 5 human annotators. (4) The Fleiss’ Kappa of GPT4 and humans over all annotations. As shown in Table \ref{tab: kappa}, the agreement between GPT-4 and humans is still higher than that among humans.

\begin{figure}[ht]
    \centering
\begin{tbox}{Context Memory}
\tiny \setlength{\columnsep}{7pt}
  \begin{multicols}{2}
% \textbf{Task:}  Context Memory \\ \\
% \vspace{10pt}
\textbf{Case:}\\ \\
\textbf{Human:} I'm looking for an interesting simulation game, do you have any recommendations? \\
\textbf{Assistant:} Absolutely! There are many fantastic simulation games out there. If you're into construction and management simulations, `Cities: Skylines' might be a good fit……\\ \\
\textbf{Human:} I would like a game that also has elements of fantasy.\\
\textbf{Assistant:} In that case, `The Elder Scrolls V: Skyrim' might be perfect for you. It is an open-world action \redtext{role-playing game} set in a fantasy realm where you can explore vast landscapes, slay monsters, complete quests, and even become a dragonborn with unique powers. Another great option is `The Witcher 3: Wild Hunt', \redtext{an epic role-playing game} with a rich fantasy world, compelling storylines, and exciting combat.

\columnbreak

\textbf{Evaluation:} \\ \\
The AI assistant's response in the last round of conversation does not demonstrate recall of previously mentioned information. The human asked for a simulation game recommendation and later specified an interest in games with elements of fantasy. \redtext{The assistant's recommendation of `The Elder Scrolls V: Skyrim' and `The Witcher 3: Wild Hunt' shifted towards action role-playing games, which are not typically categorized as simulation games. }There is no reference to the earlier part of the conversation where simulation games were the main topic. The assistant provided relevant suggestions based on the new information about the interest in fantasy but failed to integrate this with the initial request for simulation games. \\ \\
\textbf{Rating:}  [[4]]
\end{multicols}
\end{tbox}
    \caption{Case of context memory task.}
    \label{fig: case_memory}
\end{figure}
% \begin{figure}[ht]
%     \centering
%     \includegraphics[width=1.0\linewidth]{source/case_memory.pdf}
%     \caption{Case of context memory task.}
%     \label{fig: case_refer}
% \end{figure}

%%%%%%%%%%%%%%%%%%%%%%%%%%%%%%%%%%%%%%%%%%%%%%%%%%%%%%%%anaphora resolution
\begin{figure}[ht]
    \centering
    \begin{tbox}{Anaphora Resolution}
\tiny \setlength{\columnsep}{7pt}
  \begin{multicols}{2}
% \textbf{Task:}  Anaphora Resolution \\ \\
% \vspace{10pt}
\textbf{Case:}\\ \\
\textbf{Human:} What's the process for applying for a public library card?\\
\textbf{Assistant: }To apply for a library card, you need to provide a valid photo ID and proof of address. Then, you'll have to fill out an application form either online or in person at the library.\\\\
\textbf{Human: }What if there's a mistake on it?\\
\textbf{Assistant: }If there's an error on your library card, you should report it to the library staff immediately. They will help you correct the information and issue a new card.\\\\
\textbf{Human:} Is there a fee for the replacement?\\
\textbf{Assistant:} \redtext{There is usually a fee for replacing a lost or damaged library card. } The fee varies depending on the library system, but it is typically around 3 to 5.\\

\columnbreak

\textbf{Evaluation:} \\ \\
% \vspace{5pt}
The assistant's response in the last round fails to accurately reflect the referential information in the Human's question. \redtext{The Human specifically asked about a fee for replacing a card with a mistake on it, which implies a card replacement due to an error made by the library or during the production of the card. Instead, the assistant's response refers to the replacement fee for a lost or damaged card, which is not what the Human was asking about.} Therefore, the assistant did not demonstrate a correct understanding of the referential information from the Human's question and provided information that was inaccurate in the given context.\\\\
\textbf{Rating:} [[2]]
\end{multicols}
\end{tbox}
    \caption{Case of anaphora resolution task.}
    \label{fig: case_refer}
\end{figure}
% \begin{figure}[ht]
%     \centering
%     \includegraphics[width=1.0\linewidth]{source/case_refer.pdf}
%     \caption{Case of anaphora resolution task.}
%     \label{fig: case_refer}
% \end{figure}

%%%%%%%%%%%%%%%%%%%%%%%%%%%%%%%%%%%%%%%%%%%%%topic shift
\begin{figure}[ht]
    \centering
\begin{tbox}{Topic Shift}
\tiny \setlength{\columnsep}{7pt}
  \begin{multicols}{2}
% \textbf{Task:} Topic Shift \\ \\
% \vspace{10pt}
\textbf{Case:}\\ \\
\textbf{Human: }I want to watch a film with a complex plot. Any suggestions?\\
\textbf{Assistant:} For a film with a complex plot, I'd recommend `Inception' directed by Christopher Nolan. It's a thrilling science fiction heist movie that delves into the realms of dream manipulation and features several layers of narrative complexity.\\\\
\textbf{Human: }What’s a healthy breakfast option for someone on a keto diet?\\
\textbf{Assistant:} For someone on a keto diet, a healthy breakfast option could include an omelet with spinach, cheese, and avocado, which is high in healthy fats and fiber while being low in carbs. Another option is a smoothie with almond milk, spinach, peanut butter, and chia seeds.\\\\
\textbf{Human}: Does the movie leave the ending open to the audience’s interpretation?\\
\textbf{Assistant:} \redtext{The movie `Leave' doesn't exist. Could you please provide more context or clarify the title of the movie you are referring to?}\\\\

\columnbreak

\textbf{Evaluation:} \\ \\
% \vspace
The AI assistant's response in the last round of conversation fails to acknowledge the previous reference to the movie `Inception' and incorrectly assumes a misunderstanding regarding the existence of the movie title `Leave'. \redtext{This is a significant error as the human was not asking about a movie called `Leave' but inquiring about the ending of `Inception', which was discussed earlier in the conversation. The assistant's response does not detect and adapt to the continuation of the initial topic about the movie `Inception'. Instead, it seems confused, which indicates a significant lack of coherence in the dialogue. }\\
Based on the criteria provided, the response should have continued the thread about `Inception' and provided insight into the open-ended nature of the movie's conclusion. The AI's response is not relevant and does not demonstrate an understanding of the evolving conversation, which affects its coherence and the insightfulness of the answer.\\\\

\textbf{Rating:} [[1]]

\end{multicols}
\end{tbox}
    \caption{Case of topic shift task.}
    \label{fig: case_topic_switch}
\end{figure}
% \begin{figure}[ht]
%     \centering
%     \includegraphics[width=1.0\linewidth]{source/case_topic_switch.pdf}
%     \caption{Case of topic shift task.}
%     \label{fig: case_topic_switch}
% \end{figure}

%%%%%%%%%%%%%%%%%%%%%%%%%%%%%%%%%%%%%%%%%%%%content confusion
\begin{figure}[h]
    \centering
\begin{tbox}{Content Confusion}
\tiny \setlength{\columnsep}{7pt}
  \begin{multicols}{2}
% \textbf{Task:} Content Confusion \\ \\
% \vspace{10pt}
\textbf{Case:}\\ \\
% \textbf{Human:} What is ``deep learning'' in the context of artificial intelligence? \\
% \textbf{Assistant:} Deep learning is a subset of artificial intelligence (AI) that involves the use of neural networks with many layers, or "deep" networks, to model complex patterns in data. Here is a detailed explanation…… \\ \\
% \textbf{Human:} I'm looking forward to a deep learning session at the pool. Should I bring anything besides my swimsuit?\\
% \textbf{Assistant:} \redtext{Apologies for the confusion, but "deep learning" in the context of swimming refers to a different concept. In swimming, deep learning refers to a method of teaching swimming where individuals gradually progress from shallow water to deeper water.}
% So, for a deep learning session at the pool, you should bring the following besides your swimsuit……
\textbf{Human:} Could you explain the concept of `deep learning' and how it's distinct from traditional machine learning?\\
\textbf{Assistant:} Deep learning is a subset of machine learning that involves a class of algorithms known as artificial neural networks, which are designed to recognize patterns in data……\\\\
\textbf{Human:} When people mention `deep sea fishing', is it related to 'deep learning'?\\
\textbf{Assistant:} \redtext{Yes, the idea behind `deep sea fishing' is that deep learning models are designed to learn representations of data that are similar to how humans learn.} Just as deep sea fishing involves exploring the depths of the ocean to discover new species, deep learning models are designed to explore the depths of the data to discover hidden patterns and relationships.
The term `deep sea fishing' is meant to convey the idea that deep learning models are capable of discovering complex and abstract representations of data, much like how deep-sea fishing can reveal new and unexpected species in the depths of the ocean……
% In this sense, "deep sea fishing" is a metaphor for the exploratory nature of deep learning, which is designed to uncover hidden patterns and relationships in data. The term has since been widely adopted in the machine learning community as a way to describe the power and potential of deep learning models.\\\\

\columnbreak

\textbf{Evaluation:} \\ \\
The AI assistant's response in the last round of conversation fails to meet the criteria for resistance to interference and accuracy. The assistant incorrectly states that the term `deep sea fishing' is related to `deep learning'. This is not true; `deep sea fishing' is a term that refers to the activity of fishing in the deeper parts of the ocean and has nothing to do with the field of machine learning or deep learning. \\
\redtext{The assistant incorrectly introduces a non-existent relationship between the two concepts and provides an incorrect origin for the term in the context of deep learning, which demonstrates a clear influence from the previous dialogue pattern and a failure to provide accurate information based on the current question.}\\\\
\textbf{Rating:}  [[1]]
\end{multicols}
\end{tbox}
    \caption{Case of content confusion task.}
    \label{fig: case_similar}
\end{figure}
% \begin{figure}[ht]
%     \centering
%     \includegraphics[width=1.0\linewidth]{source/case_similar.pdf}
%     \caption{Case of content confusion task.}
%     \label{fig: case_similar}
% \end{figure}

%%%%%%%%%%%%%%%%%%%%%%%%%%%%%%%%%%%%%%%%content rephrasing
\begin{figure}[ht]
    \centering
\begin{tbox}{Content Rephrasing}
\tiny \setlength{\columnsep}{7pt}
  \begin{multicols}{2}
% \textbf{Task:}  Content Rephrasing \\ \\
% \vspace{10pt}
\textbf{Case:}\\ \\

\textbf{Human: }What are the characteristics that distinguish Renaissance art from other periods?\\
\textbf{Assistant:} Renaissance art is marked by a revival of classical themes and a focus on humanism, where human experience and rationality were given central importance……\\\\
\textbf{Human:} 
Could you help me convert the above information into a fun and engaging infographic description?\\
\textbf{Assistant:} 
% Unfortunately, as an AI language model, I am not able to create visual content such as infographics. However, I can provide you with the information in a text-based format that you can use to create your own infographic.\\
\redtext{
Title: Characteristics of Renaissance Art\\
1. Revival of classical themes\\
2. Focus on humanism\\
3. Realistic representation\\
4. Linear perspective\\
5. Anatomically correct figures\\
6. Naturalistic proportions\\
7. Harmony, balance, and symmetry}\\
……
% 8. Use of light and shadow\\
% 9. Rich colors and subtle gradation of tones\\
% 10. Increased use of oil paints\\
% You can use this information to create a visually appealing infographic using tools such as Canva or Adobe Spark.
\columnbreak

\textbf{Evaluation:} \\ \\
% \vspace{5pt}

The AI assistant's response does not meet the requirements set by the user. The user asked for a fun, engaging infographic description of the information, which the AI produced a simple list of characteristics. 
\redtext{The response given by the AI is a simple list of characteristics without any of the engaging or creative elements that were implied in the user's request for an infographic description.} The AI could have used more descriptive and engaging language. For example, it could have used more expressive adjectives or presented the information in a more narrative style.\\
Instead, the response is a straightforward and dry enumeration of points, which is quite the opposite of what the user asked for. The response does not align with the user's request for a format that is `fun, engaging'.\\\\

\textbf{Rating:} [[2]]
\end{multicols}
\end{tbox}
    \caption{Case of content rephrasing task.}
    \label{fig: case_rewrite_content}
\end{figure}
% \begin{figure}[ht]
%     \centering
%     \includegraphics[width=1.0\linewidth]{source/case_rewrite_content.pdf}
%     \caption{Case of content rephrasing task.}
%     \label{fig: case_rewrite_content}
% \end{figure}

%%%%%%%%%%%%%%%%%%%%%%%%%%%%%%%%%%%%%%%%%%%format rephrasing
\begin{figure}[ht]
    \centering
\begin{tbox}{Format Rephrasing}
\tiny \setlength{\columnsep}{7pt}
  \begin{multicols}{2}
% \textbf{Task:}  Format Rephrasing \\ \\
% \vspace{10pt}
\textbf{Case:}\\ \\
\textbf{Human:} What are the potential side effects of antibiotics? \\
\textbf{Assistant:} Antibiotics can have a range of potential side effects, which differ depending on the specific medication being used. Common side effects include nausea, diarrhea, and stomach pain. Some people may also experience allergic reactions……\\ \\
\textbf{Human:} Could you present the same information in a simplified list format for easier reading?\\
\textbf{Assistant:} Certainly! Here's a simplified list of potential side effects of antibiotics:\\
1. Nausea\\
2. Diarrhea\\
3. Stomach pain\\
……\\
\redtext{11. Tooth discoloration (can occur in children with certain antibiotics)\\
12. Disruption of gut flora (good bacteria in the digestive system)}

\columnbreak

\textbf{Evaluation:} \\ \\
The assistant's response maintains the original main idea, presenting the potential side effects of antibiotics in a simplified list format as requested by the human. The list includes the side effects mentioned in the original response, such as nausea, diarrhea, stomach pain, and allergic reactions.

However, \redtext{the rewritten response adds side effects that were not mentioned in the first response (vomiting, yeast infections, dizziness, headache, photo sensitivity, and tooth discoloration).} It seems that the AI assistant has expanded the list with additional information that was not part of the original content. Therefore, while the main ideas about the common side effects and the variability of individual reactions are preserved, the inclusion of new side effects that were not present in the original response fails to meet the rewriting requirement of maintaining the same content. \\ \\
\textbf{Rating:}  [[4]]
\end{multicols}
\end{tbox}
    \caption{Case of format rephrasing task.}
    \label{fig: case_rewrite_format}
\end{figure}
% \begin{figure}[ht]
%     \centering
%     \includegraphics[width=1.0\linewidth]{source/case_rewrite_format.pdf}
%     \caption{Case of format rephrasing task.}
%     \label{fig: case_rewrite_format}
% \end{figure}

%%%%%%%%%%%%%%%%%%%%%%%%%%%%%%%%%%%%%%%%%%%%%%self-correction
\begin{figure}[ht]
    \centering
\begin{tbox}{Self-Correction}
\tiny \setlength{\columnsep}{7pt}
  \begin{multicols}{2}
% \textbf{Task:}  Self-Correction \\ \\
% \vspace{10pt}
\textbf{Case:}\\ \\
\textbf{Human:} What is the most effective treatment for type 1 diabetes? \\
\textbf{Assistant:} Type 1 diabetes is best treated with dietary changes and exercise alone.\\ \\
\textbf{Human:} Are you sure about that? I believe that might not be accurate.\\
\textbf{Assistant:} \redtext{66\% confident about the statement.}

\columnbreak

\textbf{Evaluation:} \\ \\
The AI assistant's response fails to meet the evaluation criteria on multiple levels. Firstly, it does not engage with the human's query nor does it acknowledge the challenge presented by the human participant. Instead of addressing the concern raised about the accuracy of the treatment for type 1 diabetes, \redtext{the AI assistant provides a percentage of confidence in its previous statement, which is not a valid response to the human's challenge.} \\ \\
\textbf{Rating:}  [[1]]
\end{multicols}
\end{tbox}
    \caption{Case of self-correction task.}
    \label{fig: case_rethink_false}
\end{figure}
% \begin{figure}[ht]
%     \centering
%     \includegraphics[width=1.0\linewidth]{source/case_rethink_false.pdf}
%     \caption{Case of self-correction task.}
%     \label{fig: case_rethink_false}
% \end{figure}

%%%%%%%%%%%%%%%%%%%%%%%%%%%%%%%%%%%%%%%%%%%%%%%%mathematical reasoning
\begin{figure}[ht]
    \centering
\begin{tbox}{Mathematical Reasoning}
\tiny \setlength{\columnsep}{7pt}
  \begin{multicols}{2}
% \textbf{Task:}  Mathematical Reasoning \\ \\
% \vspace{10pt}
\textbf{Case:}\\ \\
\textbf{Human:} On the bookshelf, there are three different types of publications: 6 different storybooks, 5 different pictorials, and 7 different popular science books. How many different ways can Sophie choose one book from the shelf? \\
\textbf{Assistant:} Sophie can choose one book from the shelf in 18 different ways because …… \\ \\
\textbf{Human:} What if she wants to take two books of different types?\\
\textbf{Assistant:} First, we need to determine the number of choices for each type of book. Sophie can choose one storybook in 6 ways, one pictorial in 5 ways, and one popular science book in 7 ways.
\redtext{To calculate the total number of ways Sophie can choose two books of different types, we multiply the number of choices for each type:
6 ways * 5 ways * 7 ways = 210 ways.}
Therefore, Sophie can choose two books of different types in 210 different ways.

\columnbreak

\textbf{Evaluation:} \\ \\
For the second question about choosing two books of different types, the assistant's answer is incorrect. The assistant calculated the product of the number of ways to choose each type of book (6 * 5 * 7 = 210 ways), which would be the number of ways to choose one book from each type if Sophie were to take three books. \redtext{However, Sophie is only choosing two books of different types, not one from each type.} The correct approach, as given in the reference solution, is to calculate the sum of the products of the numbers of books from two different types at a time: (6 * 5) + (6 * 7) + (5 * 7) = 30 + 42 + 35 = 107 ways. \redtext{Therefore, the assistant's reasoning process for the second question is fundamentally flawed, as it does not correctly understand the requirement of choosing only two books, not three.}
 \\ \\
\textbf{Rating:}  [[5]]
\end{multicols}
\end{tbox}
    \caption{Case of mathematical reasoning task.}
    \label{fig: ase_infer_math}
\end{figure}
% \begin{figure}[ht]
%     \centering
%     \includegraphics[width=1.0\linewidth]{source/case_infer_math.pdf}
%     \caption{Case of mathematical reasoning task.}
%     \label{fig: ase_infer_math}
% \end{figure}

%%%%%%%%%%%%%%%%%%%%%%%%%%%%%%%%%%%%%%%%%%%%%general reasoning
\begin{figure}[ht]
    \centering
\begin{tbox}{General Reasoning}
\tiny \setlength{\columnsep}{7pt}
  \begin{multicols}{2}
% \textbf{Task:}  General Reasoning \\ \\
% \vspace{10pt}
\textbf{Case:}\\ \\
\textbf{Human:} I have five boxes A, B, C, D and E. The weight of box A is less than B and B is less than C. Can you identify the heaviest box so far? \\
\textbf{Assistant:} Based on the provided details, C is the heaviest box among A, B, and C. \\ \\
\textbf{Human:} Now we have two more boxes, F and G. F weighs more than B and G weighs more than F. Can you tell me which is the heaviest box now?\\
\textbf{Assistant:} From the original information, you indicated that C is heavier than both A and B. You then added that F is heavier than B, and G is heavier than F. Therefore, \redtext{G is the heaviest box among all the mentioned boxes} (A, B, C, D, E, F, and G), since no information about D and E affecting the order of weight has been provided.

\columnbreak

\textbf{Evaluation:} \\ \\
\redtext{The AI assistant made a logical leap by determining G as the heaviest without having direct comparisons between C, F, and G.} The reasoning process was generally correct except for this last deduction, which did not consider the lack of comparison between the weights of boxes C, D, E, and G. The assistant should have acknowledged that without direct comparisons involving C, D, E, and G, one cannot conclude which is the heaviest overall. 

The communication was clear and the reasoning was mostly logical, yet it missed the critical step of recognizing the absence of necessary comparative information to make a final determination. \\ \\
\textbf{Rating:}  [[3]]
\end{multicols}
\end{tbox}
    \caption{Case of general reasoning task.}
    \label{fig: case_infer_gener}
\end{figure}
% \begin{figure}[ht]
%     \centering
%     \includegraphics[width=1.0\linewidth]{source/case_infer_gener.pdf}
%     \caption{Case of general reasoning task.}
%     \label{fig: case_infer_gener}
% \end{figure}
\clearpage
\newpage
%%%%%%%%%%%%%%%%%%%%%%%%%%%%%%%%%%%%instruction clarification
\begin{figure}[ht]
    \centering
\begin{tbox}{Instruction Clarification}
\tiny \setlength{\columnsep}{7pt}
  \begin{multicols}{2}
% \textbf{Task:}  Instruction Clarification \\ \\
% \vspace{10pt}
\textbf{Case:}\\ \\
\textbf{Human:} What's the best book to read? \\
\textbf{Assistant:} \redtext{Here are some highly regarded books across various genres:}\\
1. `To Kill a Mockingbird' by Harper Lee\\
2. `1984' by George Orwell\\
……\\

It is recommended to explore different genres and authors to discover what resonates with your personal taste.

\columnbreak

\textbf{Evaluation:} \\ \\
In this scenario, using a counter-question would have been beneficial to ascertain more about the human's preferences to provide a more tailored recommendation. \redtext{The response does offer a variety of options, which could be useful considering the broad nature of the question. However, without further clarification, it's impossible to determine if any of these recommendations align with the human's preferences.} Additionally, the assistant does not attempt to seek further information from the human that would allow for a more personalized suggestion.
 \\ \\
\textbf{Rating:}  [[4]]
\end{multicols}
\end{tbox}
    \caption{Case of instruction clarification task.}
    \label{fig: case_asking}
\end{figure}
% \begin{figure}[ht]
%     \centering
%     \includegraphics[width=1.0\linewidth]{source/case_asking.pdf}
%     \caption{Case of instruction clarification task.}
%     \label{fig: case_asking}
% \end{figure}

%%%%%%%%%%%%%%%%%%%%%%%%%%%%proactive interaction
\begin{figure}[ht]
    \centering
\begin{tbox}{Proactive Interaction}
\tiny \setlength{\columnsep}{7pt}
  \begin{multicols}{2}
% \textbf{Task:}  Proactive Interaction  \\ \\
% \vspace{10pt}
\textbf{Case:}\\ \\
\textbf{Human:} I couldn't believe how windy it was today. It nearly blew me away! \\
\textbf{Assistant:} \redtext{The gusts were so strong that I struggled to walk straight, and my hair was constantly whipped around my face.} The wind carried with it a chill that ……

\columnbreak

\textbf{Evaluation:} \\ \\
The AI assistant's reply is exceptionally descriptive and imaginative, creating a vivid portrayal of the experience in windy conditions. However, for the purpose of this evaluation, which is focused on interactivity, the response is less effective.
\redtext{While the assistant's narrative is indeed engaging, it is more of a monologue than an interactive dialogue. It does not build on the human's initial comment with the aim of fostering a back-and-forth exchange.} Therefore, the response does not excel in the proactive initiation and sustenance of an engaging conversation. \\ \\
\textbf{Rating:}  [[3]]
\end{multicols}
\end{tbox}
    \caption{Case of proactive interaction task.}
    \label{fig: case_interact}
\end{figure}
% \begin{figure}[ht]
%     \centering
%     \includegraphics[width=1.0\linewidth]{source/case_interact.pdf}
%     \caption{Case of proactive interaction task.}
%     \label{fig: case_interact}
% \end{figure}

\end{document}